\documentclass[letterpaper]{article} 
\usepackage{aaai24}  
\usepackage{times}  
\usepackage{helvet}  
\usepackage{courier}  
\usepackage[hyphens]{url}  
\usepackage{graphicx} 
\usepackage{natbib}  
\usepackage{caption} 
\usepackage{algorithm}
\usepackage{algorithmic, appendix}
\usepackage{newfloat}
\usepackage{listings}
\DeclareCaptionStyle{ruled}{labelfont=normalfont,labelsep=colon,strut=off} 
\floatstyle{ruled}
\newfloat{listing}{tb}{lst}{}
\floatname{listing}{Listing}
\usepackage{amsmath}
\usepackage{amssymb}
\usepackage{mathtools}

\usepackage[textsize=tiny]{todonotes}

\usepackage{subfigure}
\usepackage{enumitem}
\usepackage{gensymb}
\usepackage{svg}
\usepackage{multirow}
\usepackage{fixltx2e} 
\usepackage{tikz-cd}
\usepackage{scrextend}
\usepackage[new]{old-arrows}
\usepackage{booktabs}
\usepackage{amsmath}
\usepackage{amssymb}
\usepackage{mathtools}
\usepackage{amsthm}
\usepackage[capitalize,noabbrev]{cleveref}
\usepackage[textsize=tiny]{todonotes}

\newtheorem{theorem}{Theorem}
\newtheorem{remark}[theorem]{Remark}

\newcommand{\sehookarrow}{\mathrel{\rotatebox[origin=c]{-45}{$\hookrightarrow$}}} 
\newcommand{\swhookarrow}{\mathrel{\rotatebox[origin=c]{-135}{$\hookrightarrow$}}} 

\newcommand{\wh}{\widehat}
\newcommand{\wt}{\widetilde}
\newcommand{\G}{\mathcal{G}}
\newcommand{\h}{\mathcal{H}}
\newcommand{\I}{\mathcal{I}}
\newcommand{\X}{\mathcal{X}}
\newcommand{\V}{\mathcal{V}}
\newcommand{\E}{\mathcal{E}}
\newcommand{\W}{\mathcal{W}}
\newcommand{\M}{\mathbf{M}}
\newcommand{\R}{\mathbb{R}}
\newcommand{\e}{\epsilon}

\title{Time-Aware Knowledge Representations of\\ Dynamic Objects with Multidimensional Persistence}
\author {
    Baris Coskunuzer\equalcontrib\textsuperscript{\rm 1},
    Ignacio Segovia-Dominguez\equalcontrib\textsuperscript{\rm 2},
    Yuzhou Chen\equalcontrib\textsuperscript{\rm 3},
    Yulia R. Gel\textsuperscript{\rm 1,4}\\
}
\affiliations {
    \textsuperscript{\rm 1}University of Texas at Dallas, Department of Mathematical Sciences\\  
    \textsuperscript{\rm 2}West Virginia University, School of Mathematical \& Data Sciences\\
    \textsuperscript{\rm 3}Temple University, Department of Computer and Information Sciences\\   
    \textsuperscript{\rm 4}National Science Foundation\\
   coskunuz@utdallas.edu, Ignacio.SegoviaDominguez@mail.wvu.edu, yuzhou.chen@temple.edu, ygl@utdallas.edu
}

\begin{document}

\maketitle

\begin{abstract}
Learning time-evolving objects such as multivariate time series and dynamic networks requires the development of novel knowledge representation mechanisms and neural network architectures, which allow for capturing implicit time-dependent information contained in the data. Such information is typically not directly observed but plays a key role in the learning task performance. In turn, lack of time dimension in knowledge encoding mechanisms for time-dependent data leads to frequent model updates, poor learning performance, and, as a result, subpar decision-making.  Here we propose a new approach to a time-aware knowledge representation mechanism that notably focuses on implicit time-dependent topological information along multiple geometric dimensions. In particular, we propose a new approach, named \textit{Temporal MultiPersistence} (TMP), which produces multidimensional topological fingerprints of the data by using the existing single parameter topological summaries. The main idea behind TMP is to merge the two newest directions in topological representation learning, that is, multi-persistence which simultaneously describes data shape evolution along multiple key parameters, and zigzag persistence to enable us to extract the most salient data shape information over time.    
We derive theoretical guarantees of TMP vectorizations and show its utility, in application to forecasting on benchmark traffic flow, Ethereum blockchain, and electrocardiogram datasets, demonstrating the competitive performance, especially, in scenarios of limited data records. In addition, our TMP method improves the computational efficiency of the state-of-the-art multipersistence summaries up to 59.5 times.
\end{abstract}

\section{Introduction} \label{sec:intro}



Over the last decade, the field of topological data analysis (TDA) has demonstrated its effectiveness in revealing concealed patterns within diverse types of data that conventional methods struggle to access. Notably, in cases where conventional approaches frequently falter, tools such as persistent homology (PH) within TDA have showcased remarkable capabilities in identifying both localized and overarching patterns. These tools have the potential to generate a distinctive topological signature, a trait that holds great promise for a range of ML applications. This inherent capacity of PH becomes particularly appealing for capturing implicit temporal traits of evolving data, which might hold the crucial insights underlying the performance of learning tasks.


In turn, the concept of multiparameter persistence (MP) introduces a groundbreaking dimension to machine learning by enhancing the capabilities of persistent homology. Its objective is to analyze data across multiple dimensions concurrently, in a more nuanced manner. However, due to the complex algebraic challenges intrinsic to its framework, MP has yet to be universally defined in all contexts~\cite{botnan2022introduction,carriere2020multiparameter}. 

In response, we present a novel approach designed to effectively harness MP homology for the dual purposes of time-aware learning and the representation of time-dependent data. Specifically, the temporal parameter within time-dependent data furnishes the crucial dimension necessary for the application of the slicing concept within the MP framework. Our method yields a distinctive topological MP signature for the provided time-dependent data, manifested as multidimensional vectors (matrices or tensors). These vectors are highly compatible with ML applications. Notably, our findings possess broad applicability and can be tailored to various forms of PH vectorization, rendering them suitable for diverse categories of time-dependent data.

\smallskip

{\bf Our key contributions} can be summarized as follows:

\begin{itemize}

    \item We bring a new perspective to use TDA for time-dependent data by using multipersistence approach. 
    
    \item We introduce TMP vectorizations framework which provides a multidimensional topological fingerprint of the data. TMP framework expands many known single persistence vectorizations to multidimensions by utilizing time dimension effectively in PH machinery.
    
    \item The versatility of our TMP framework allows its application to diverse types of time-dependent data.
    Furthermore, we show that TMP enjoys many important stability guarantees as most single persistence summaries.
    
    \item Rooted in computational linear algebra, TMP vectorizations generate multidimensional arrays (i.e., matrices or tensors) 
    which serve as compatible inputs for various ML models. 
    Notably, our proposed TMP approach boasts a speed advantage, performing up to 59.5 times faster than the cutting-edge MP methods.
    
    \item Through successful integration of the latest TDA techniques with deep learning tools, our TMP-Nets model consistently and cohesively outperforms the majority of state-of-the-art deep learning models.
    
\end{itemize}






\section{Related Work} \label{sec:relatedwork}

   

\subsection{Time Series Forecasting} \label{sec:timeseries_rel}
Recurrent Neural Networks (RNNs) are the most successful deep learning techniques to model datasets with time-dependent variables~\cite{lipton2015critical}. Long-Short-Term Memory networks (LSTMs) addressed the prior RNN limitations in learning long-term dependencies by solving known issues with exploding and vanishing gradients~\cite{RRN:Yong:2019}, serving as basis for other improved RNN, such as Gate Recurrent Units (GRUs)~\cite{GRU:Dey:2017}, Bidirectional LSTMs (BI\_LSTM)~\cite{BiLSTM:Wang:2018}, and seq2seq LSTMs~\cite{seq2seqLSTM:Sutskever:2014}. Despite the widespread adoption of RNNs in multiple applications~\cite{Rainfall:Xiang:2020,LSTMcompanies,GRU:Shin:2020,shewalkar2019performance,segovia2021tlife,bin2018describing}, RNNs are limited by the structure of the input data and can not naturally handle data-structures from manifolds and graphs, i.e. non-Euclidean spaces.    



\subsection{Graph Convolutional Networks} \label{sec:GCN_rel}
New methods on graph convolution-based methods overcome prior limitations of traditional GCN approaches, e.g. learning underlying local and global connectivity patterns~\cite{velivckovic2017graph,DefferrardNIPS,kipf2016semi}. GCN handles graph-structure data via aggregation of node information from the neighborhoods using graph filters. Lately, there is an increasing interest in expanding GCN capabilities to the time series forecasting domain. In this context, modern approaches have reached outstanding results in COVID-19 forecasting, money laundering, transportation forecasting, and scene recognition~\cite{pareja2020evolvegcn,ignacioCOVID19KDD,yu2018spatio,yan2018spatial,guo2019attention,weber2019anti,yao2018deep}. However, a major drawback of these approaches is the lack of versatility as they assume a fixed graph-structure and rely on the existing correlation among spatial and temporal features. 


\subsection{Multiparameter Persistence} \label{sec:MP_rel}
Multipersistence (MP) is a highly promising approach to significantly improve the success of single parameter persistence (SP) in applied TDA, but the theory is not complete yet \citep{botnan2022introduction}. 
Except for some special cases, the MP theory tends to suffer from the problem of the nonexistence of barcode decomposition because of the partially ordered structure of the index set $\{(\alpha_i,\beta_j)\}$.  
The existing approaches remedy this issue via the slicing technique by studying one-dimensional fibers of the multiparameter domain. 
However, this approach tends to lose most of the information the MP approach produces. Another idea along these lines is to use several such directions (vineyards), and produce a vectorization summarizing these SP vectorizations \citep{carriere2020multiparameter}. However, again choosing these directions suitably and computing restricted SP vectorizations are computationally costly which restricts these approaches in many real-life applications. There are several promising recent studies in this direction~\cite{botnan2021signed,vipond2020multiparameter}, but these techniques often do not provide a topological summary that can readily serve as input to ML models.
In this paper, we develop a highly efficient way to use the MP approach for time-dependent data and provide a multidimensional topological summary with TMP Vectorizations. We discuss the current fundamental challenges in the MP theory and the contributions of our TMP vectorizations in~\Cref{sec:MP_theory}.

\section{Background} \label{sec:background}


We start by providing the basic background for our machinery. While our techniques are applicable to any type of time-dependent data, here we mainly focus on the dynamic networks since our primary motivation comes from time-aware learning of time-evolving graphs as well as time series and spatio-temporal processes, also represented as graph structures. 
(For discussion on other types of data see \Cref{sec:othertypedata}.) 

\noindent {\em Notation Table:} All the notations used in the paper are given in \cref{notations} in the appendix.

\smallskip

\noindent \textit{Time-Dependent Data:} 
Throughout the paper, by \textit{time-dependent data}, we mean the data which implicitly or explicitly has time information embedded in itself. Such data include but are not limited to multivariate time series, spatio-temporal processes, and dynamic networks. Since our paper is primarily motivated by time-aware graph neural networks and their broader applications to forecasting, we focus on dynamic networks.  
Let $\{\mathcal{G}_1, \mathcal{G}_2, \dots, \mathcal{G}_{T}\}$ be a sequence of weighted graphs for time steps $t= \{1,\ldots, T\}$. In particular, $\mathcal{G}_t = \{\mathcal{V}_t, \mathcal{E}_t, W_t$\} with node set $\mathcal{V}_t$, and edge set $\mathcal{E}_t$. Let $|\mathcal{V}_t|=N_t$ be the cardinality of the node set. $W_t$ represents the edge weights for $\mathcal{E}_t$ as a nonnegative symmetric $N_t\times N_t$-matrix with entries $\{\omega^t_{rs}\}_{1\leq r,s\leq N_t}$, i.e. the adjacency matrix of $\mathcal{G}_t$. In other words, $\omega^t_{rs} > 0$ for any $e^t_{rs} \in \mathcal{E}_t$ and $\omega^t_{rs} = 0$, otherwise. In the case of unweighted networks, let $\omega^t_{rs} =1$ for any $e^t_{rs} \in \mathcal{E}_t$ and $\omega^t_{rs} = 0$, otherwise.

\subsection{Background on Persistent Homology} \label{sec:PH_background}
Persistent homology (PH) is a mathematical machinery to capture the hidden shape patterns in the data by using algebraic topology tools. PH extracts this information by keeping track of the evolution of the topological features (components, loops, cavities) created in the data while looking at it using different resolutions. Here, we give basic background for PH in the graph setting. For further details, see~\cite{dey2022computational,edelsbrunner2010computational}. 

For a given graph $\mathcal{G}$, consider a nested sequence of subgraphs $\mathcal{G}^{1} \subseteq \ldots \subseteq \mathcal{G}^{N}=\mathcal{G}$. For each $\mathcal{G}^i$, define an abstract simplicial complex 
$\wh{\mathcal{G}}^{i}$, $1\leq i\leq N$, yielding a filtration of complexes $\wh{\mathcal{G}}^{1} \subseteq \ldots \subseteq \wh{\mathcal{G}}^{N}$. 
Here, clique complexes are among the most common ones, i.e., clique complex $\wh{\G}$ is obtained by assigning (filling with) a $k$-simplex to each complete $(k+1)$-complete subgraph in $\G$, e.g., a $3$-clique, a complete $3$-subgraph, in $\G$ will be filled with a $2$-simplex (triangle). Then, in this sequence of simplicial complexes, one can systematically keep track of the evolution of the topological patterns mentioned above. A $k$-dimensional topological feature (or $k$-hole) may represent connected components ($0$-hole), loops ($1$-hole) and cavities ($2$-hole). For each $k$-hole $\sigma$, PH records its first appearance in the filtration sequence, say $\wh{\mathcal{G}}^{b_\sigma}$, and first disappearance in later complexes, $\wh{\mathcal{G}}^{d_\sigma}$ 
with a unique pair $(b_\sigma, d_\sigma)$, where $1\leq b_\sigma<d_\sigma\leq N$
We call $b_\sigma$ \textit{the birth time} of $\sigma$ and $d_\sigma$ \textit{the death time} of $\sigma$. We call $d_\sigma-b_\sigma$ \textit{the life span} of $\sigma$. 
PH records all these birth and death times of the topological features in \textit{persistence diagrams}. Let $0\leq k\leq D$ where $D$ is the highest dimension in the simplicial complex $\wh{\G}^N$. Then $k^{th}$ persistence diagram ${\rm{PD}_k}(\G)=\{(b_\sigma, d_\sigma) \mid \sigma\in H_k(\wh{\mathcal{G}}^i) \mbox{ for } b_\sigma\leq i<d_\sigma\}$. Here, $H_k(\wh{\mathcal{G}}^i)$ represents the $k^{th}$ homology group of $\wh{\mathcal{G}}^i$ which keeps the information of the $k$-holes in the simplicial complex $\wh{\mathcal{G}}^i$. With the intuition that the topological features with a long life span (persistent features) describe the hidden shape patterns in the data, these persistence diagrams provide a unique topological fingerprint of $\mathcal{G}$.



As one can easily notice, the most important step in the PH machinery is the construction of the nested sequence of subgraphs $\mathcal{G}^{1} \subseteq \ldots \subseteq \mathcal{G}^{N}=\mathcal{G}$. For a given unweighted graph $\G=(\V,E)$, the most common technique is to use a filtering function $f:\mathcal{V}\to\R$ with a choice of thresholds $\I=\{\alpha_i\}_1^N$ where $\alpha_1=\min_{v \in \V} f(v)<\alpha_2<\ldots<\alpha_N=\max_{v \in \V} f(v)$. For $\alpha_i\in \I$, let $\V_i=\{v_r\in\V\mid f(v_r)\leq \alpha_i\}$. Let $\G^i$ be the induced subgraph of $\G$ by $\V_i$, i.e. $\G^i=(\V_i,\E_i)$ where $\E_i=\{e_{rs}\in \E\mid v_r,v_s\in\V_i\}$. This process yields a nested sequence of subgraphs $\G^1\subset \G^2\subset \ldots \subset\G^N=\G$, called \textit{the sublevel filtration} induced by the filtering function $f$. Choice of $f$ is crucial here, and in most cases, $f$ is either an important function from the domain of the data, e.g. amount of transactions or volume transfer,  
or a function defined from intrinsic properties of the graph, e.g. degree, betweenness. Similarly, for a weighted graph, one can use sublevel filtration on the weights of the edges and obtain a suitable filtration reflecting the domain information stored in the edge weights.  For further details on different filtration types of networks, see~\cite{aktas2019persistence,hofer2020graph}. 

\subsection{Multidimensional Persistence} \label{sec:MP_background}
In the previous section, we discussed the single-parameter persistence theory. The reason for the term "single" is that we filter the data in only one direction $\G^1\subset \dots\subset\G^N=\G$. Here, the choice of direction 
is the key to extracting the hidden patterns from the observed data.
For some tasks and data types, it is sufficient to consider only one dimension (or filtering
function $f:\V\to\R$) (e.g., atomic numbers for protein networks) in order to extract the intrinsic data properties. 
However, often the observed data
may have more than one direction to be analyzed (for example, in the case of money laundering detection on bitcoin, we may need to use both transaction amounts and numbers of transactions between any two traders). 
With this intuition, multiparameter persistence (MP) theory is suggested as a natural generalization of single persistence (SP).


In simpler terms, if one uses only one filtering function, sublevel sets induce a single parameter filtration $\wh{\G}^1\subset \dots\subset\wh{\G}^N=\wh{\G}$. Instead, if one uses two or more functions, then it would enable us to study finer substructures and patterns in the data. 
In particular, let $f:\V\to\R$ and $g:\V\to\R$ be two filtering functions with very valuable complementary information of the network. Then, MP idea is presumed to produce a unique topological fingerprint combining the information from both functions. These pair of functions  $f,g$ induces a multivariate filtering function $F:\V \mapsto \R^2$ with $F(v)=(f(v),g(v))$. Again, we can define a set of nondecreasing thresholds $\{\alpha_i\}_1^m$ and $\{\beta_j\}_1^n$ for $f$ and $g$ respectively. Then, $\V^{ij}=\{v_r\in V\mid f(v_r)\leq \alpha_i , g(v_r)\leq\beta_j\}$, i.e. $\V^{ij}=F^{-1}((-\infty,\alpha_i]\times(-\infty,\beta_j])$. Then, let $\G^{ij}$ be the induced subgraph of $\G$ by $\V^{ij}$, i.e. the smallest subgraph of $\G$ containing $\V^{ij}$. Then, 
instead of a single filtration of complexes, we get 
a bifiltration of complexes $\{\wh{\G}^{ij}\mid 1\leq i\leq m, 1\leq j\leq n\}$.  See \Cref{Fig:ToyMP} (Appendix) for an explicit example.

As illustrated in \cref{Fig:ToyMP}, we can imagine $\{\wh{\G}^{ij}\}$ as a rectangular grid of size $m\times n$  such that for each $1\leq i_0\leq m$, $\{\wh{G}^{i_0j}\}_{j=1}^n$ gives a nested (horizontal) sequence of simplicial complexes. Similarly, for each $1\leq j_0\leq n$, $\{\wh{G}^{ij_0}\}_{i=1}^m$ gives a nested (vertical) sequence of simplicial complexes. 
By computing the homology groups of these complexes, $\{H_k(\G^{ij})\}$, we obtain the induced bigraded persistence module (a rectangular grid of size $m\times n$). 
Again, the idea is to keep track of the $k$-dimensional topological features via the homology groups $\{H_k(\wh{\G}^{ij})\}$ in this grid. As detailed in \Cref{sec:MP_theory}, because of the technical issues related to commutative algebra coming from the partially ordered structure of the multipersistence module, this MP approach has not been completed like SP theory yet, and there is a need to facilitate this promising idea effectively in real-life applications.


In this paper, for time-dependent data, we overcome this problem by using the naturally inherited special direction in the data: Time. By using this canonical direction in the multipersistence module, we bypass the partial ordering problem and generalize the ideas from single parameter persistence to produce a unique topological fingerprint of the data via MP.  Our approach provides a general framework to utilize various vectorization forms defined for single PH and gives a multidimensional topological summary of the data. 

\noindent \textit{Utilizing Time Direction - Zigzag Persistence: } 
While our intuition is to use time direction in MP for forecasting purposes, the time parameter is not very suitable to use in PH construction in its original form. This is because PH construction needs nested subgraphs to keep track of the existing topological features, while time-dependent data do not come nested, i.e. $\G_{t_1}\not\subseteq \G_{t_2}$ in general for $t_1\leq t_2$. However, a generalized version of PH construction helps us to overcome this problem. We want to keep track of topological features which exist in different time instances. Zigzag homology~\cite{carlsson2010zigzag} bypasses the requirement of the nested sequence by using the "zigzag scheme". We provide the details of zigzag persistent homology in \cref{sec:zigzag}.

\section{TMP Vectorizations} \label{sec:TMP}

We now introduce a general framework to define vectorizations for multipersistence homology on time-dependent data. First, we recall the single persistence vectorizations which we will expand as multidimensional vectorizations with our TMP framework.



\subsection{Single Persistence Vectorizations} \label{sec:SPvec}
While PH extracts hidden shape patterns from data as persistence diagrams (PD), PDs being a collection of points $\{(b_\sigma,d_\sigma)\}$ in $\R^2$ by itself are not very practical for statistical and machine learning purposes. Instead, the common techniques are by accurately representing PDs as kernels~\cite{kriege2020survey} or vectorizations~\cite{ali2022survey}. \textit{Single Persistence Vectorizations} transform obtained PH information (PDs) into a function or a feature vector form which are much more suitable for ML tools. Common single persistence (SP) vectorization methods are Persistence Images~\cite{adams2017persistence}, Persistence Landscapes~\cite{Bubenik:2015}, Silhouettes~\cite{chazal2014stochastic}, and various Persistence Curves~\cite{chung2019persistence}. These vectorizations define a single variable or multivariable functions out of PDs, which can be used as fixed size $1D$ or $2D$ vectors in applications, i.e $1\times m$ vectors or $m\times n$ vectors. For example, a Betti curve for a PD with $m$ thresholds can be written as $1\times m$ size vectors. Similarly, persistence images is an example of $2D$ vectors with the chosen resolution (grid) size. See the examples below and in \cref{sec:MPVexamples} for further details.

\subsection{TMP Vectorizations} \label{sec:TMP_vec}
Finally, we define our Temporal MultiPersistence (TMP) framework for time-dependent data. In particular, by using the existing single-parameter persistence vectorizations, we produce multidimensional vectorization by effectively using the time direction in the multipersistence module. The idea is to use zigzag homology in the time direction and consider $d$-dimensional filtering for the other directions. This process produces $(d+1)$-dimensional vectorizations of the dataset. While the most common choice would be $d=1$ for computational purposes, we restrict ourselves to $d=2$ to give a general idea. The construction can easily be generalized to higher dimensions. Below and in \cref{sec:MPVexamples}, we provide explicit examples of TMP Vectorizations. While we mainly focus on network data in this part, we give how to generalize TMP vectorizations to other types of data (e.g., point clouds, images) in \Cref{sec:othertypedata}.

Again, let $\wt{\G}=\{\mathcal{G}_1, \mathcal{G}_2, \dots, \mathcal{G}_{T}\}$  be a sequence of weighted (or unweighted) graphs for time steps $t=1,\ldots, T$ with $\mathcal{G}_t = \{\mathcal{V}_t, \mathcal{E}_t, W_t$\} as defined in~\cref{sec:background}. By using a filtering function $F_t:\V_t\to\R^d$ or weights, define a bifiltration for each $t_0$, i.e. $\{\G_{t_0}^{ij}\}$ for $1\leq i\leq m$ and $1\leq j\leq n$. For each fixed $i_0,j_0$, consider the sequence $\{\mathcal{G}^{i_0j_0}_1, \mathcal{G}_2^{i_0j_0}, \dots, \mathcal{G}_{T}^{i_0j_0}\}$. This sequence of subgraphs induces a zigzag sequence of clique complexes as described in \cref{sec:zigzag}:
\begin{eqnarray*}
\wh{\G}_1^{i_0j_0}\hookrightarrow\wh{\G}^{i_0j_0}_{1.5}\hookleftarrow\wh{\G}^{i_0j_0}_2\hookrightarrow\wh{\G}^{i_0j_0}_{2.5}\hookleftarrow\wh{\G}_3\hookrightarrow\dots\hookleftarrow\wh{\G}^{i_0j_0}_T.
\end{eqnarray*}

Now, let $ZPD_k(\wt{\G}^{i_0j_0})$ be the induced zigzag persistence diagram. 
Let $\varphi$ represent an SP vectorization as described above, e.g. Persistence Landscape, Silhouette, Persistence Image, Persistence Curves. This means if $PD(\G)$ is the persistence diagram for some filtration induced by $\G$, then we call $\varphi(\G)$ is the corresponding vectorization for $PD(\G)$ (see Figure \ref{Fig:MPZigzag} in Appendix~F7). 
In most cases, $\varphi(\G)$ is represented as functions on the threshold domain (Persistence curves, Landscapes, Silhouettes, Persistence Surfaces). However, the discrete structure of the threshold domain enables us to interpret the function $\varphi(\G)$ as a $1D$-vector $\vec{\varphi}(\G)$  (Persistence curves, Landscapes, Silhouettes) or $2D$-vector $\vec{\varphi}(\G)$  (Persistence Images). See examples given below and in the \cref{sec:MPVexamples} for more details.

\begin{figure}[t!]
    \centering
    \includegraphics[width=0.46\textwidth]{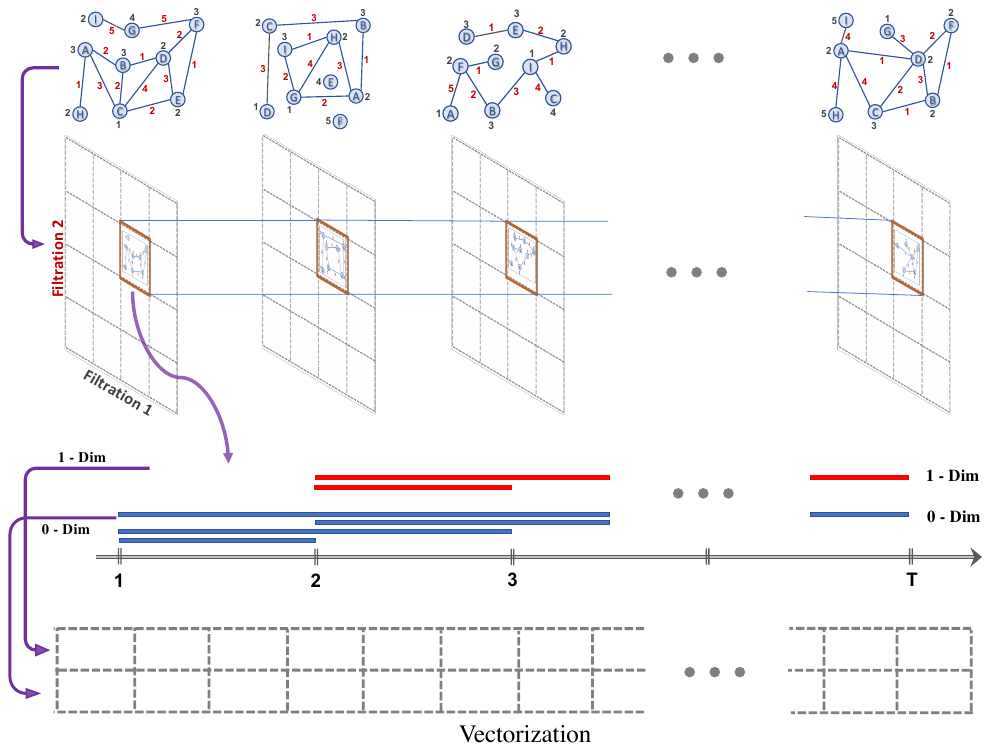}
    \caption{\footnotesize 
    TMP outline. 
    Given 
    $\wt{\G}=\{\mathcal{G}_1, \mathcal{G}_2, \dots, \mathcal{G}_{T}\}$ with time-index $t=1,\ldots, T$ (1st row) we apply a bifiltration on node/edge-features at $t$, i.e. $\{\G_{t}^{ij}\}$ for $1\leq i\leq m$ and $1\leq j\leq n$ (2nd row). The sequence of subgraphs $\{\mathcal{G}^{i_0j_0}_1, \mathcal{G}_2^{i_0j_0}, \dots, \mathcal{G}_{T}^{i_0j_0}\}$, at fixed $i_0,j_0$ is the input into the zigzag persistence method to produce a zigzag persistence barcode (3rd row). Then, $\vec{\varphi}(\wt{\G}^{i_0j_0})$ is the corresponding vectorization for zigzag PD
    $ZPD_k(\wt{\G}^{i_0j_0})$ of $k-$dim feature (4th row). 
    \label{Fig:MPZigzag}}
    \vspace{-.1in}
\end{figure}

Now, let $\vec{\varphi}(\wt{\G}^{ij})$ be the corresponding vector for the zigzag persistence diagram $ZPD_k(\wt{\G}^{ij})$. Then, for any $1\leq i\leq m$ and $1\leq j\leq n$, we have a ($1D$ or $2D$) vector $\vec{\varphi}(\wt{\G}^{ij})$. Now, define the induced TMP Vectorization $\M_\varphi$ as the corresponding tensor $\M_\varphi^{ij}=\Vec{\varphi}(\wt{\G}^{ij})$ for $1\leq i\leq m$ and $1\leq j\leq n. $

In particular, if $\vec{\varphi}$ is a $1D$-vector of size $1\times k$, then $\M_\varphi$ would be a $3D$-vector (rank-$3$ tensor) with size $m\times n \times k$. if $\vec{\varphi}$ is a $2D$-vector of size $k\times r$, then $\M_\varphi$ would be a rank-$4$ tensor with size $m\times n \times k\times r$. In the examples below, we provide explicit constructions for $\M_\varphi$ for the most common SP vectorizations $\varphi$.

\subsection{Examples of TMP Vectorizations} \label{sec:examples}
While we describe TMP Vectorizations for $d=2$, 
in most applications, $d=1$ would be preferable for computational purposes. Then if the preferred single persistence (SP) vectorization $\varphi$ produces $1D$-vector (say size $1\times r$), then induced TMP Vectorization would be a $2D$-vector $M_\varphi$ (a matrix) of size $m\times r$ where $m$ is the number of thresholds for the filtering function used, e.g. $f:\V_t\to \R$. These $m\times r$ matrices provide unique topological fingerprints for each time-dependent dataset $\{\G_t\}_{t=1}^T$. These multidimensional fingerprints are produced by using persistent homology with two-dimensional filtering where the first dimension is the natural direction time $t$, and the second dimension comes from the filtering function $f$. 

Here, we discuss explicit constructions of two examples of TMP vectorizations. As we mentioned above, the framework is very general, and it can be applied to various vectorization methods. In \cref{sec:MPVexamples}, we provide details of further examples of TMP Vectorizations for time-dependent data, i.e., TMP Silhouettes, and TMP Betti Summaries.


\subsubsection{TMP Landscapes} 

Persistence Landscapes $\lambda$ are one of the most common SP vectorization methods introduced in~\cite{Bubenik:2015}. For a given persistence diagram $PD(\G)=\{(b_i,d_i)\}$, $\lambda$ produces a function $\lambda(\G)$ by using generating functions $\Lambda_i$ for each $(b_i,d_i)\in PD(\G)$, i.e. $\Lambda_i:[b_i,d_i]\to\R$ is a piecewise linear function obtained by two line segments starting from $(b_i,0)$ and $(d_i,0)$ connecting to the same point $(\frac{b_i+d_i}{2},\frac{b_i-d_i}{2})$. Then, the \textit{Persistence Landscape} function $\lambda(\G):[\e_1,\e_q]\to\R$ is defined as $\lambda(\G)(t)=\max_i\Lambda_i(t)$ for $t\in [\e_1,\e_q]$, where $\{\e_k\}_1^q$ represents the thresholds for the filtration used.

Considering the piecewise linear structure of the function, $\lambda(\G)$ is completely determined by its values on $2q-1$ points, i.e. $\frac{b_i\pm d_i}{2}\in\{\e_1, \e_{1.5}, \e_2, \e_{2.5}, \dots ,\e_q\}$ where $\e_{k.5}={(\e_k+\e_{k+1})}/{2}$. Hence, a vector of size $1\times (2q-1)$ whose entries the values of this function would suffice to capture all the information needed, i.e.
$\vec{\lambda}=[ \lambda(\e_1)\ \lambda(\e_{1.5})\ \lambda(\e_2)\ \lambda(\e_{2.5})\ \lambda(\e_3)\ \dots \ \lambda(\e_q)]$.

Now, for the time-dependent data $\wt{\G}=\{\G_t\}_{t=1}^T$, to construct our induced TMP Vectorization $\M_\lambda$, TMP Landscapes, we use $\lambda$ for time direction, $t=1,\dots, T$. For zigzag persistence, we have $2T-1$ thresholds steps. Hence, by taking $q=2T-1$, we would have $4T-3$ length vector $\vec{\lambda}(\wt{\G})$.

For the other multipersistence direction, by using a filtering function $f:\V_t\to \R$ with the threshold set $\I=\{\alpha_j\}_1^m$, we obtain \textit{TMP Landscape} $\M_\lambda$ as follows:
$\M_\lambda^j=\vec{\lambda}(\wt{\G}^j)$ where $\M_\lambda^j$ represents $j^{th}$-row of the $2D$-vector $\M_\lambda$. Here, $\wt{\G}^j=\{\G_t^j\}_{t=1}^T$ is induced by the sublevel filtration for $f:\V_t\to\R$ as described in the paper, i.e. $\G^j_t$ is the induced subgraph by $\V^j_t=\{v_r\in \V_t\mid f(v_r)\leq \alpha_j\}$. 

Hence, for a time-dependent data $\wt{\G}=\{\G_t\}_{t=1}^T$, TMP Landscape $\M_\lambda(\wt{\G})$ is a $2D$-vector of size $m\times (4T-3)$ where $T$ is the number of time steps.


\subsubsection{TMP Persistence Images} 

Next SP vectorization in our list is persistence images~\cite{adams2017persistence}. Different than most SP vectorizations, persistence images produce $2D$-vectors. The idea is to capture the location of the points in the persistence diagrams with a multivariable function by using the $2D$ Gaussian functions centered at these points. For $PD(\G)=\{(b_i,d_i)\}$, let $\phi_i$ represent a $2D$-Gaussian centered at the point $(b_i,d_i)\in \R^2$. Then, one defines a multivariable function, \textit{Persistence Surface}, $\wt{\mu}=\sum_iw_i\phi_i$ where $w_i$ is the weight, mostly a function of the life span $d_i-b_i$. To represent this multivariable function as a $2D$-vector, one defines a $k\times l$ grid (resolution size) on the domain of $\wt{\mu}$, i.e. threshold domain of $PD(\G)$. Then, one obtains the \textit{Persistence Image}, a $2D$-vector $\vec{\mu}=[\mu_{rs}]$  of size $k\times l$, where $\mu_{rs}=\int_{\Delta_{rs}}\wt{\mu}(x,y)\,dxdy$ and $\Delta_{rs}$ is the corresponding pixel (rectangle) in the $k\times l$ grid.

Following a similar route, for our TMP vectorization, we use time as one direction, and the filtering function in the other direction, i.e. $f:\V_t\to \R$ with threshold set $\I=\{\alpha_j\}_1^m$. Then, for time-dependent data $\wt{\G}=\{\G_t\}_{t=1}^T$, in the time direction, we use zigzag PDs and their persistence images. Hence, for each $1\leq j\leq m$, we define \textit{TMP Persistence Image} as $\M_\mu^j(\wt{\G})=\vec{\mu}(\wt{\G}^j)$ where $2D$-vector $\M_\mu^j$ is $j^{th}$-floor of the $3D$-vector $\M_\mu$. Then, TMP Persistence Image $\M_\mu^j(\wt{\G})$ is a $3D$-vector of size $m\times k\times l$. 

More details for TMP Persistence Surfaces and TMP Silhouettes are provided in \Cref{sec:MPVexamples}.

\subsection{Stability of TMP Vectorizations}  \label{sec:stability}
We now prove that when the source single parameter vectorization $\varphi$ is stable, then so is its induced TMP vectorization $\M_\varphi$. We discuss the details of the stability notion in persistence theory and examples of stable SP vectorizations in \cref{sec:stability2}.

Let $\wt{\G}=\{\G_t\}_{t=1}^T$ and $\wt{\h}=\{\h_t\}_{t=1}^T$ be two time sequences of networks. Let $\varphi$ be a stable SP vectorization with the stability equation  
$$\mathrm{d}(\varphi(\wt{\G}),\varphi(\wt{\h}))\leq C_\varphi\cdot \W_{p_\varphi}(PD(\wt{\G}),PD(\wt{\h}))$$
for some $1\leq p_\varphi\leq \infty$. Here, $\W_p$ represents Wasserstein-$p$ distance as defined in \cref{sec:stability2}.

Now, consider the bifiltrations $\{\wh{\G}_t^{ij}\}$ and $\{\wh{\h}_t^{ij}\}$ for each $1\leq t\leq T$. We define the induced matching distance between the multiple persistence diagrams (See \cref{rmk:matching_distance}) as 
{\small $\mathbf{D}(\{ZPD(\wt{\G})\},\{ZPD(\wt{\h})\})=\max_{i,j}\W_{p_\varphi}(ZPD(\wt{\G}^{ij}), ZPD(\wt{\h}^{ij}))$}

Now, define the distance between TMP Vectorizations as $\mathbf{D}(\M_\varphi(\wt{\G}),\M_\varphi(\wt{\h}))=\max_{i,j} \mathrm{d}(\varphi(\wt{\G}^{ij}),\varphi(\wt{\h}^{ij}))$.

\begin{theorem} \label{thm:stability} Let $\varphi$ be a stable vectorization for single parameter PDs. Then, the induced TMP Vectorization $\M_\varphi$ is also stable, i.e. With the notation above, there exists $\wh{C}_\varphi>0$ such that for any pair of time-aware network sequences $\wt{\G}$ and $\wt{\h}$, we have the following inequality.
$$\mathbf{D}(\M_\varphi(\wt{\G}),\M_\varphi(\wt{H}))\leq \wh{C}_\varphi\cdot \mathbf{D}(\{ZPD(\wt{\G})\},\{ZPD(\wt{\h})\})$$

\end{theorem} 

The proof of the theorem is given in \cref{sec:stability3}.


\nocite{langley00}


\section{TMP-Nets} \label{sec:TMP}

To fully take advantage of the extracted signatures by TMP vectorizations, we propose a GNN-based module to track and learn significant temporal and topological patterns. Our Time Aware Multiparameter Persistence Nets (TMP-Nets) capture spatio-temporal relationships via trainable node embedding dictionaries in a GDL-based framework.

\subsection*{Graph Convolution on Adaptive Adjacency Matrix} \label{sec:matrix}
To model the hidden dependencies among nodes in the spatio-temporal graph, we define the spatial graph convolution operation based on the adaptive adjacency matrix and given node feature matrix. Inspired by~\cite{wu2019graph}, to investigate the beyond pairwise relations among nodes, we use the adaptive adjacency matrix based on trainable node embedding dictionaries, i.e.,
    $Z^{(\ell)}_{t, \text{Spatial}} = L Z_{t, \text{Spatial}}^{(\ell-1)}W^{(\ell-1)}$,
where $L = \text{Softmax}(\text{ReLU}(E_{\theta}E^{\top}_{\theta}))$ (here $E_{\theta} \in \mathbb{R}^{N \times d_c}$ and $d_c \geq 1$), $ Z_{\text{Spatial}}^{(\ell-1)}$ and $ Z_{\text{Spatial}}^{(\ell)}$ are the input and output of $(\ell-1)$-th layer, and $ Z_{\text{Spatial}}^{(0)} = X \in \mathbb{R}^{N \times F}$ (here $F$ represents the number of features for each node), and $W^{(\ell-1)}$ is the trainable weights.

\subsection*{Topological Signatures Representation Learning} \label{sec:Rep_learning}
In our experiments, we use CNN based model to learn the TMP topological features. Given the TMP topological features of resolution $p$, i.e., $\text{TMP}_t \in \mathbb{R}^{p \times p}$, we employ CNN-based model and global max pooling to obtain the image-level local topological feature $Z_{t, \text{TMP}}$ as
    $$Z_{t, \text{TMP}} = f_{\text{GMP}}(f_{\theta}(\text{TMP}_t)),$$
where $f_{\text{GMP}}$ is the global max pooling, $f_{\theta}$ is a CNN based neural network with parameter set $\theta_i$, and $Z_{t, \text{TMP}} \in \mathbb{R}^{d_c}$ is the output for TMP representation.

Lastly, we combine the two embeddings to obtain the final embedding $Z_{t}$:
    $$Z_t = Concat(Z_{t, \text{Spatial}}, Z_{t, \text{TMP}}).$$
To capture both spatial and temporal correlations in time-series, we feed the final embedding $Z_t$ into Gated Recurrent Units (GRU) for future time points forecasting.

\section{Experiments}\label{sec:Experiments}
\subsubsection{Datasets:} We consider three types of data: two widely used benchmark datasets on California (CA) \textbf{traffic}~\cite{chen2001freeway} and
\textbf{electrocardiography} (ECG5000)~\cite{dau2019ucr}, and the newly emerged data on Ethereum \textbf{blockchain} tokens~\cite{shamsi2022chartalist}.
(The results on the ECG5000 are presented in \Cref{sec:ECG5000}). More details descriptions of datasets can be found in \Cref{sec:datasets}.

\subsection{Experimental Results}
We compare our TMP-Nets with 6 state-of-the-art baselines.
We use three standard performance metrics 
Mean Absolute Error (MAE), Root Mean Square Error (RMSE), and Mean absolute percentage error (MAPE). 
We provide additional experimental results in \Cref{sec:Other_Results}. 
In \cref{sec:Experimental_Setup_TMP}, we provide further details on the experimental setup and empirical evaluation. Our source code is available at the link~\footnote{\url{https://www.dropbox.com/sh/h28f1cf98t9xmzj/AACBavvHc_ctCB1FVQNyf-XRa?dl=0}}.

\begin{table*}[t]
\centering
\resizebox{.7\linewidth}{!}{
\begin{tabular}{lccc}
\toprule
\textbf{{Model}} & \textbf{{Bytom}} & \textbf{{Decentraland}} & \textbf{{Golem}} \\
\midrule
{DCRNN}~\citep{li2018diffusion}& {35.36$\pm$1.18} & {27.69$\pm$1.77} & {23.15$\pm$1.91}\\
{STGCN}~\citep{yu2018spatio}& {37.33$\pm$1.06} & {28.22$\pm$1.69} & {23.68$\pm$2.31} \\
{GraphWaveNet}~\citep{wu2019graph} &{39.18$\pm$0.96} &{37.67$\pm$1.76} & {28.89$\pm$2.34} \\
{AGCRN}~\citep{bai2020adaptive} &{34.46$\pm$1.37} & {26.75$\pm$1.51} & {22.83$\pm$1.91}\\
{Z-GCNETs}~\cite{chen2021z} & \underline{31.04$\pm$0.78} & \underline{23.81$\pm$2.43} & \textbf{22.32$\pm$1.42}\\
{StemGNN}~\citep{cao2020spectral} &{34.91$\pm$1.04} &{28.37$\pm$1.96} & \underline{22.50$\pm$2.01}\\
\midrule
\textbf{TMP-Nets} &\textbf{28.77$\pm$3.30} &\textbf{22.97$\pm$1.80} &29.01$\pm$1.05 \\
\bottomrule
\end{tabular}}
\caption{Experimental results on Bytom, Decentraland, and Golem \textcolor{black}{on MAPE and standard deviation}.\label{prediction_results}}
\end{table*}

\smallskip

\noindent {\em Results on Blockchain Datasets:} Table~\ref{prediction_results} shows performance on Bytom, Decentraland, and Golem.
Table~\ref{prediction_results} suggests the following phenomena: (i) TMP-Nets achieves the best performance on Bytom and Decentraland, and the relative gains of TMP-Nets over the best baseline (i.e., Z-GCNETs) are 7.89\% and 3.66\% on Bytom and Decentraland respectively; (ii) compared with Z-GCNETs, the size of TMP topological features used in this work is much smaller than the zigzag persistence image utilized in Z-GCNETs. 

An interesting question is why TMP-Nets performs differently on Golem vs. Bytom and Decentraland. Success on each network token depends on the diversity of connections among nodes. In cryptocurrency networks, we expect nodes/addresses to be connected with other nodes with similar transaction connectivity (e.g. interaction among whales) as well as with nodes with low connectivity (e.g. terminal nodes). However, the assortativity measure of Golem (-0.47) is considerably lower than Bytom (-0.42) and Decentraland (-0.35), leading to disassortativity patterns (i.e., repetitive isolated clusters) in the Golem network, which, in turn, downgrade the success rate of forecasting.  

\begin{table*}[t]
\centering
\setlength\tabcolsep{4pt}
\begin{tabular}{lccc|ccc}
\toprule
\multirow{2}{*}{\textbf{Model}}& \multicolumn{3}{c}{\textbf{PeMSD4}} & \multicolumn{3}{c}{\textbf{ PeMSD8}}
\\
\cmidrule(lr){2-4}\cmidrule(lr){5-7}
           & MAE & RMSE & MAPE {(\%)} & MAE & RMSE & MAPE {(\%)}\\
\midrule
AGCRN&\underline{110.36$\pm$0.20} &150.37$\pm$0.15 &\underline{208.36$\pm$0.20} &87.12$\pm$0.25 &109.20$\pm$0.33 &277.44$\pm$0.26\\
Z-GCNETs&112.65$\pm$0.12&153.47$\pm$0.17&{\bf 206.09$\pm$0.33} &69.82$\pm$0.16 &95.83$\pm$0.37 &{\bf 102.74$\pm$0.53}\\
StemGNN&112.83$\pm$0.07 &\underline{150.22$\pm$0.30} &209.52$\pm$0.51 &\underline{65.16$\pm$0.36} &\underline{89.60$\pm$0.60} &\underline{108.71$\pm$0.51}\\
\midrule
\textbf{TMP-Nets} &{\bf 108.38$\pm$0.10} & {\bf 147.57$\pm$0.23} &208.66$\pm$0.27   &{\bf 59.82$\pm$0.82} &{\bf 85.86$\pm$0.64} &109.88$\pm$0.65 \\
\bottomrule
\end{tabular}
\caption{Forecasting performance on (first 1,000 networks) of PeMSD4 and PeMSD8 benchmark datasets.
\label{traffic_forecast_result_1}}
\end{table*}

\begin{table*}[h!]
\centering
\setlength\tabcolsep{4pt}
\begin{tabular}{lccc|ccc}
\toprule
\multirow{2}{*}{\textbf{Model}}& \multicolumn{3}{c}{\textbf{PeMSD4}} & \multicolumn{3}{c}{\textbf{ PeMSD8}}
\\
\cmidrule(lr){2-4}\cmidrule(lr){5-7}
           & MAE & RMSE & MAPE {(\%)} & MAE & RMSE & MAPE {(\%)}\\
\midrule
AGCRN&90.36$\pm$0.10 & 122.61$\pm$0.13 & 176.90$\pm$0.35 & 55.20$\pm$0.19 & 83.01$\pm$0.53 & 167.39$\pm$0.25\\
Z-GCNETs&\underline{89.57$\pm$0.11} & \underline{117.94$\pm$0.15} & \underline{180.11$\pm$0.26}& 47.11$\pm$0.20 & \underline{80.25$\pm$0.24} & \underline{98.15$\pm$0.33}\\
StemGNN&93.27$\pm$0.16 & 131.49$\pm$0.21 & 189.18$\pm$0.30 & \underline{53.86$\pm$0.39} & 82.00$\pm$0.52 & {\bf 97.78$\pm$0.30}\\
\midrule
\textbf{TMP-Nets} &{\bf 85.15$\pm$0.12} & {\bf 115.00$\pm$0.16} & {\bf 170.97$\pm$0.22} & {\bf 50.20$\pm$0.37} & {\bf 80.17$\pm$0.26} & 100.31$\pm$0.58\\
\bottomrule
\end{tabular}
\caption{Forecasting performance on (first 2,000 networks) PeMSD4 and PeMSD8 benchmark datasets.
\label{traffic_forecast_result_2}}
\vspace{-3mm}
\end{table*}


\smallskip

\noindent {\em Results on Traffic Datasets:} For traffic flow data PeMSD4 and PeMSD8, we evaluate Z-GCNETs' performance on varying lengths. This allows us to further explore the learning capabilities of our Z-GCNETs as a function of sample size. In particular,
in many real-world scenarios, there exists only a limited number of temporal records to be used in the training stage, and the learning problem with lower sample sizes becomes substantially more challenging.  
Tables~\ref{traffic_forecast_result_1} and~\ref{traffic_forecast_result_2} show that under the scenario of limited data records for both PeMSD4 and PeMSD8 (i.e., $\mathcal{T} = 1,000$ and $\mathcal{T}^\prime = 2,000$), our TMP-Nets always outperforms three representative baselines in MAE and RMSE. For example, TMP-Nets significantly outperform SOTA baselines, where we achieve relative gains of 1.79\% and 4.36\% in RMSE on PeMSD4$_{\mathcal{T} = 1,000}$ and PeMSD8$_{\mathcal{T} = 1,000}$, respectively. Overall, the results demonstrate that our proposed TMP-Nets can accurately capture the hidden complex spatial and temporal correlations in the correlated time series datasets and achieve promising forecasting performances under the scenarios of limited data records. Moreover, we conduct experiments on the whole PeMSD4 and PeMSD8 datasets. As Table~\ref{traffic_forecast_result_3} (Appendix) indicates, our TMP-Nets still achieve competitive performances on both datasets. 

Finally, we applied our approach in a different domain with a benchmark electrocardiogram dataset, ECG5000. Again, our model gives highly competitive results with the SOTA methods (\Cref{sec:ECG5000}).

\subsubsection{Ablation Studies:}
To better evaluate the importance of different components of TMP-Nets, we perform ablation studies on two traffic datasets, i.e., PeMSD4 and PeMSD8 by using only (i) $Z^{(\ell)}_{t, \text{Spatial}}$ or (ii) $Z_{t, \text{TMP}}$ as input.  \Cref{tab:ablation_Zspatial_PEMSD4} 
report the forecasting performance of (i) $Z^{(\ell)}_{t, \text{Spatial}}$, (ii) $Z_{t, \text{TMP}}$, and (iii) TMP-Nets (our proposed model). We find that our TMP-Nets outperforms both $Z^{(\ell)}_{t, \text{Spatial}}$ and $Z_{t, \text{TMP}}$ on two datasets, yielding highly statistically significant gains. Hence, we can conclude that (i) TMP vectorizations help to better capture global and local hidden topological information in the time dimension, and (ii) spatial graph convolution operation accurately learns the inter-dependencies (i.e., spatial correlations) among spatio-temporal graphs. We provide further ablation studies comparing the effect of slicing direction and the MP vectorization methods in the \Cref{sec:ablation_appendix}. 


\begin{table}[t]
\centering
\begin{tabular}{lc|c}
\toprule
\textbf{{Model}} &\textbf{{PeMSD4}} &  \textbf{{PeMSD8}}\\
\midrule
TMP-Nets & {\bf 147.57}$\pm${\bf 0.23}  & {\bf 85.86}$\pm${\bf 0.64}\\ 
$Z_{t, \text{TMP}}$ & 165.67$\pm$0.30 & 90.23$\pm$0.15   \\
$Z^{(\ell)}_{t, \text{Spatial}}$ & 153.75$\pm$0.22 & 88.38$\pm$1.05  \\
\bottomrule
\end{tabular}
\caption{
 Ablation Study on PeMSD4 and PESMD8 (RMSE results for first 1000 networks). 
\label{tab:ablation_Zspatial_PEMSD4}}
\vspace{-.1in}
\end{table}

\subsubsection{Computational Complexity:} One of the key issues why MP has not propagated widely into practice yet is its high computational costs. 
Our method improves on the state-of-the-art MP (ranging from 23.8 to 59.5 times faster than Multiparameter Persistence Image (MP-I)~\cite{carriere2020multiparameter}, and from 1.2 to 8.6 times faster than Multiparameter Persistence Kernel (MP-L)~\cite{corbet2019kernel}) and, armed with a computationally fast vectorization method (e.g., Betti summary~\cite{lesnick2022computing}),  TMP yields competitive computational costs for a lower number of filtering functions (See~\Cref{sec:computationaltime}). Nevertheless, scaling for really large scale-problems is still a challenge. In the future we will explore TMP constructed only on the landmark points, that is, TMP will be constructed not on all but on the most important {\it landmark} nodes, which would lead to substantial sparsification of the graph representation. 


\subsubsection{Comparison with Other Topological GNN Models for Dynamic Networks:}
\label{sec:compTMP}

The two existing time-aware 
topological GNNs for dynamic networks are TAMP-S2GCNets~\cite{chen2021tamp} and Z-GCNETs~\cite{chen2021z}. The pivotal distinction between our model and these counterparts lies in the fact that our model serves as a comprehensive extension of both, applicable across diverse data types encompassing point clouds and images (see Section \ref{sec:othertypedata}). Z-GCNETs employs single persistence approach, rendering it unsuitable for datasets that encompass two or more significant domain functions. In contrast, TAMP-S2GCNets employs multipersistence; however, its Euler-Characteristic surface vectorization fails to encapsulate lifespan information present in persistence diagrams. Notably, in scenarios involving sparse data,   barcodes with longer lifespans signify main data characteristics, while short barcodes are considered as topological noise. The limitation of Euler-Characteristic Surfaces, being simply a linear combination of bigraded Betti numbers, lies in its inability to capture this distinction. In stark contrast, our framework encompasses all forms of vectorizations, permitting practitioners to choose their preferred vectorization technique while adapting to dynamic networks or time-dependent data comprehensively. For instance, compared to TAMP-S2GCNets model, our TMP-Nets achieves a better performance on the Bytom dataset, i.e., TMP-Nets (MAPE: 28.77$\pm$3.30) vs. TAMP-S2GCNets (MAPE: 29.26$\pm$1.06). Furthermore, from the computational time perspective, the average computation time of TMP and Dynamic Euler-Poincar\'e Surface (which is used in TAMP-S2GCNets model) are 1.85 seconds and 38.99 seconds respectively, i.e., our TMP is more efficient.

\section{Discussion}\label{sec:Discussion}

We have proposed a new highly computationally efficient summary for multidimensional persistence for time-dependent objects,  Temporal MultiPersistence (TMP). By successfully combining the latest TDA methods with deep learning tools, our TMP approach outperforms many popular state-of-the-art deep learning models in a consistent and unified manner. 
Further, we have shown that TMP enjoys important theoretical stability guarantees. As such, TMP makes an important step toward bringing the theoretical concepts of multipersistence from pure mathematics to the machine learning community and to the practical problems of time-aware learning of time-conditioned objects, such as dynamic graphs, time series, and spatio-temporal processes.  

Still, scaling for ultra high-dimensional processes, especially in modern data streaming scenarios, may be infeasible for TMP. In the future, we will investigate algorithms such as those based on landmarks or pruning, with the goal to advance the computational efficiency of TMP for streaming applications. 

\section*{Acknowledgements}
Supported by the NSF grants  DMS-2220613, DMS-2229417, ECCS 2039701, TIP-2333703,
Simons Foundation grant \# 579977, and ONR grant 
N00014-21-1-2530. Also, the paper is based upon work supported by (while Y.R.G. was serving at) the NSF. The views expressed in the article do not necessarily represent the views of NSF or ONR.

\bibliography{ZigzagGCNETs_ref,refV4}


\clearpage

\setcounter{page}{1}

\appendix

\centerline{\Large \bf Appendix}

\ 

In this part, we give additional details about our experiments and methods. In \Cref{sec:Other_Results}, we provide more experimental results as ablation studies and additional baselines. In \Cref{sec:Experimental_Setup_TMP}, we discuss datasets and our experimental setup. In \Cref{Sec:MorePH}, we provide a more theoretical background on persistent homology. In \Cref{sec:moreTMP}, we give further examples of TMP vectorizations and generalizations to general types of data. We also discuss fundamental challenges in applications of multipersistence theory in spatio-temporal data, and our contributions in this context in \Cref{sec:MP_theory}. Finally, in \Cref{sec:stability3}, we prove the stability of TMP vectorizations. \textbf{Our notation table} (\Cref{notations}) can be found at the end of the appendix.

\section{Additional Results on Experiments}\label{sec:Other_Results}

\subsection{Additional Baselines}


 Contrary to other papers (e.g., \cite{Survey:Jiang:2022}) which consider only a single option of 16,992 (PeMSD4) / 17,856 (PeMSD8) time stamps, we evaluate Z-GCNETs performance on varying lengths of 1,000 and 2,000 (Section 6.3 - Experimental Results details). This allows us to further explore the learning capabilities of our Z-GCNETs as a function of sample size and also most importantly to assess the performance of Z-GCNETs and its competitors under a more challenging and much more realistic scenario of limited temporal records. To better highlight the effectiveness of our proposed TMP-Nets model, we compare it with more baselines - DCRNN~\cite{li2018diffusion}, STGCN~\cite{yu2018spatio}, and GraphWaveNet~\cite{wu2019graph}. As shown in Table~\ref{tab:baselines_PEMSD4}, we can find that our TMP-Nets is highly statistically significantly better than DCRNN, STGCN, and GraphWaveNet on PeMSD4 dataset.

\begin{table}[H]
\centering
\begin{tabular}{lcc}
\toprule
\textbf{{Dataset}} & \textbf{{Model}} &\textbf{{RMSE}} \\
\midrule
PEMSD4 & TMP-Nets & \textbf{147.57$\pm$0.23}  \\ 
PEMSD4 & DCRNN & 153.34$\pm$0.55  \\
PEMSD4 & STGCN & 174.75$\pm$0.35  \\
PEMSD4 & GraphWaveNet & 151.87$\pm$0.22  \\
\bottomrule
\end{tabular}
\caption{Comparison of TMP-Nets and baselines on PeMSD4 (first 1,000 networks). \label{tab:baselines_PEMSD4}}
\end{table}

\begin{table}[h!]
\centering
\setlength\tabcolsep{2pt}
\resizebox{0.46\textwidth}{!}{
\begin{tabular}{lccc|ccc}
\toprule
\multirow{2}{*}{\textbf{Model}}& \multicolumn{3}{c}{\textbf{PeMSD4}} & \multicolumn{3}{c}{\textbf{ PeMSD8}}
\\
\cmidrule(lr){2-4}\cmidrule(lr){5-7}
           & MAE & RMSE & MAPE {(\%)} & MAE & RMSE & MAPE {(\%)}\\
\midrule
AGCRN& 19.83 & 32.26 & 12.97\% & 15.95 & 25.22 &10.09\%\\
Z-GCNETs& {\bf 19.50} & {\bf 31.61} & {\bf 12.78\%} & {\bf 15.76} & {\bf 25.11} & {\bf 10.01\%}\\
StemGNN& 20.24 & 32.15 & 10.03\% & \underline{15.83} & \underline{24.93} & \underline{9.26\%}\\
\midrule
\textbf{TMP-Nets} & \underline{19.57} & \underline{31.69} & \underline{12.89\%} & 16.36 & 25.85 & 10.36\%\\
\bottomrule
\end{tabular}}
\caption{Forecasting performance on whole PeMSD4 and PeMSD8 datasets.
\label{traffic_forecast_result_3}}
\end{table}

\subsection{Further Ablation Studies} \label{sec:ablation_appendix}



\subsubsection{Slicing Direction.}
To investigate the importance of time direction, we now consider zigzag persistent homology along the axis of degree instead of time. We then conduct comparison experiments between TMP-Nets (i.e., $Z_{t, \text{TMP}}$ is generated through the time axis) and TMP$_{deg}$-Nets (i.e., $Z_{t, \text{TMP}}$ is generated along the axis of degree instead of time). As Table \ref{tab:ablation_shuffle} indicates, TMP-Nets based on the time component outperforms TMP$_{deg}$-Nets. These findings can be expected, as time is one of the core variables in spatio-temporal processes, and, hence, we can conclude that extracting the zigzag-based topological summary along the time dimension is important for forecasting tasks. Nevertheless, we would like to underline that the TMP idea can be also applied to non-time-varying processes as long as there exists some alternative natural geometric dimension.

\begin{table}[H]
\centering
\begin{tabular}{lcc}
\toprule
\textbf{{Dataset}} & \textbf{{Model}} &\textbf{{MAPE}} \\
\midrule
Bytom & TMP-Nets & \textbf{28.77$\pm$3.30}  \\ 
Bytom & TMP$_{deg}$-Nets & 29.15$\pm$4.17  \\
\bottomrule
\end{tabular}
\caption{Comparison of TMP-Nets and TMP$_{deg}$-Nets on Bytom dataset.\label{tab:ablation_shuffle}}
\end{table}

\subsubsection{Bigraded Betti Numbers vs. TMP.} To compare the effectiveness of $Z_{t, \text{TMP}}$ which facilitates time direction with zigzag persistence for spatio-temporal forecasting, we conduct additional experiments on traffic datasets, i.e., PeMSD4 and PeMSD8 by using (i) TMP-Nets (based on Z-Meta) and (ii) MP$_{Betti}$-Nets (based on bigraded Betti numbers).
Tables \ref{tab:ablation_grid_PEMSD4} below show the results when using bigraded Betti numbers as a source of topological signatures in the ML model.  As Tables 3 and 4 indicate, our TMP-Nets achieves better forecasting accuracy (i.e., lower RMSE) on both PeMSD4 and PeMSD8 datasets than the MP$_{Betti}$-Nets and the difference in performance is highly statistically significant. 

Such results can be potentially attributed to the fact that  TMP-Nets tends to better capture the most important topological signals by choosing a suitable vectorization method for the task at hand. In particular, MP$_{Betti}$-Nets only counts the number of topological features but do not give any emphasis to the longer barcodes appearing in the temporal direction, that is, MP$_{Betti}$-Nets are limited in distinguishing topological signals from topological noise.  However, longer barcodes (or density of the short barcodes) in the temporal dimension typically are the key to accurately capturing intrinsic topological patterns in the spatio-temporal data.


\begin{table}[H]
\centering
\begin{tabular}{lcc}
\toprule
\textbf{{Model}} &\textbf{{PEMSD4}}&\textbf{{PEMSD8}}  \\
\midrule
TMP-Nets & \textbf{147.57$\pm$0.23 } & \textbf{85.86$\pm$0.64} \\ 
MP$_{Betti}$-Nets & 151.58$\pm$0.19  & 87.71$\pm$0.70  \\
\bottomrule
\end{tabular}
\caption{TMP-Nets vs. MP$_{Betti}$-Nets on PeMSD4 and PeMSD8 (RMSE results for first 1,000 networks). \label{tab:ablation_grid_PEMSD4}}
\end{table}



\begin{table*}[h!]
\centering
\setlength{\tabcolsep}{5pt}
\resizebox{.8\linewidth}{!}{
\begin{tabular}{lcccccc}
\toprule
\textbf{{Dataset}} & \textbf{{Dim}} &\textbf{{Betweenness}} & \textbf{{Closeness}} & \textbf{{Degree}} & \textbf{{Power-Tran}} & \textbf{{Power-Volume}} \\
\midrule
Bytom & \{0,1\} & 236.95 seg & 239.36 seg & 237.60 seg & 987.90 seg & 941.39 seg \\ 
Decentraland & \{0,1\} & 134.75 seg & 138.81 seg & 133.82 seg & 2007.50 seg & 1524.10 seg \\
Golem & \{0,1\} & 571.35 seg & 581.36 seg & 573.93 seg & 4410.47 seg & 4663.52 seg \\

\bottomrule
\end{tabular}}
\caption{Computational time on Ethereum tokens using five different filtering functions.\label{computational_time_ethereum}}
\end{table*}

\subsection{Computational Time on Different Filtering Functions} \label{sec:computationaltime}

As expected, Table~\ref{computational_time_ethereum} shows that computational time highly depends on the complexity of the selected filtering function. However, the time spent in computing TMP Vectorizations is below two hours at most, which makes our approach highly useful in ML tasks. 

\subsection{Experiments on ECG5000 Benchmark Dataset} \label{sec:ECG5000}


To support that our methodology can be applied in other dynamic networks, we run additional experiments in the ECG5000 dataset, \cite{chen2015general}. This benchmark dataset contains 140 nodes and 5,000 time stamps. When running our methodology we extract patterns via Betti, Silhouette, and Entropy vectorizations, set the window size to 12, and the forecasting step as 3. Following preliminary cross-validation experiments, we set the resolution to 50 where we use a quantile-based selection of thresholds. We perform edge-weight filtration on graphs created via a correlation matrix.   In our experiments, we have found that there is no significant difference between the results based on Betti and Silhouette vectorizations. In Table~\ref{tab:exp_ECG5000}, we only report the results of TMP-Nets based on Silhouette vectorization. From Table~\ref{tab:exp_ECG5000}, we can find that our TMP-Nets either deliver on par or outperforms the state-of-the-art baselines (with a smaller standard deviation). Note that ECG5000 is a small dataset (that is, 140 nodes only), and as such, the differences among models cannot be expected to be high for such a smaller network. 

\begin{table}[H]
\centering
\begin{tabular}{lcc}
\toprule
\textbf{{Dataset}} & \textbf{{Model}} &\textbf{{RMSE}} \\
\midrule
ECG5000 & TMP-Nets & 0.52$\pm$0.005  \\ 
ECG5000 & StemGNN & 0.52$\pm$0.006  \\
ECG5000 & AGCRN & 0.52$\pm$0.008  \\
ECG5000 & DCRNN & 0.55$\pm$0.005  \\
\bottomrule
\end{tabular}
\caption{Comparison of TMP-Nets and baselines on ECG5000.\label{tab:exp_ECG5000}}
\end{table}


\subsection{Connectivity in Ethereum networks} \label{Sec:StatEthereum}

An interesting question is why TMP-Nets performs differently on Golem vs. Bytom and Decentraland. Success on each network token depends on the diversity of connections among nodes. In cryptocurrency networks, we expect nodes/addresses to be connected with other nodes with similar transaction connectivity (e.g. interaction among whales) as well as with nodes with low connectivity (e.g. terminal nodes). However, the assortativity measure of Golem (-0.47) is considerably lower than Bytom (-0.42) and Decentraland (-0.35), leading to disassortativity patterns (i.e., repetitive isolated clusters) in the Golem network, which, in turn, downgrade the success rate of forecasting.

\begin{table}[H]
\centering
\resizebox{.95\linewidth}{!}{
\begin{tabular}{lcccc}
\toprule
 \textbf{Token} & \textbf{Degree} & \textbf{Betweenness} & \textbf{Density} & \textbf{Assortativity}   \\
 \midrule 
 Bytom & 0.1995789 & 0.0002146 & 0.0020159 &   -0.4276000   \\
 Decentraland & 0.3387378 & 0.0004677 & 0.0034215 &  -0.3589580    \\
 Golem & 0.3354401 & 0.0004175 & 0.0033882 & -0.4731063  \\ 
 \bottomrule
\end{tabular}}
\caption{Comparison of statistics on Ethereum token networks.\label{tab:Ethereum}}
\end{table}


\section{Further Details on Experimental Setup}\label{sec:Experimental_Setup_TMP}

\subsection{Datasets} \label{sec:datasets}

\subsubsection{CA Traffic.} We consider two traffic flow datasets, i.e., PeMSD4 and PeMSD8, in California from January 1, 2018, to February 28, 2018, and from January 7, 2016, to August 31, 2016, respectively. Note that, both PeMSD4 and PeMSD8 are aggregated to 5 minutes, which means there are 12 time points in the flow data per hour. Following the settings of~\cite{guo2019attention}, we split the traffic datasets with ratio $6: 2: 2$ into training, validation, and test sets; furthermore, in our experiments, we evaluate our TMP-Nets and baselines on two traffic flow datasets with varying lengths of sequences, i.e., $\mathcal{T} = 1,000$ (first 1,000 networks out of whole dataset) and $\mathcal{T}^\prime = 2,000$ (first 2,000 networks out of whole dataset).

\subsubsection{Electrocardiogram.} We use the electrocardiogram (ECG5000) dataset (i.e., with a length of 5,000) from the UCR time series archive~\cite{dau2019ucr}, where each time series length is 140.

\subsubsection{Ethereum blockchain tokens.} We use three token networks from the Ethereum blockchain (Bytom, Decentraland and Golem) each with more than \$100M in market value (\url{https://EtherScan.io}).
Thus dynamic networks are a compound of addresses of users, i.e. nodes, and daily transactions among users, i.e. edges~\cite{Tokens:Angelo:2020,akcora2021blockchain}. Since original token networks have an average of 442788/1192722 nodes/edges, we compute a subgraph via a maximum weight subgraph approximation~\cite{MaxSubgraph:Vassilevska:2006} using the amount of transactions as weight. The dynamic networks contain different numbers of nets since every token was created on different days; Bytom (285), Decentraland (206), and Golem (443). Hence, given the dynamic network $\mathcal{G}_t = \{\mathcal{V}_t, \mathcal{E}_t,\Tilde{W}_t\}$ and its corresponding node feature matrix $X_t \in \mathbb{R}^{N\times F}$, where $F$ represents the number of features, we test our algorithm with both node and edge features and use the set of more active nodes, i.e. $N=100$. 


\subsection{Experimental Setup} We implement our TMP-Nets with Pytorch framework on NVIDIA GeForce RTX 3090 GPU. Further, for all datasets, TMP-Nets is trained end-to-end by using the Adam optimizer with a L1 loss function. For Ethereum blockchain token networks, we use Adam optimizer with weight decay, initial learning rate, batch size, and epoch as 0, 0.01, 8, 80 respectively. For traffic datasets, we use Adam optimizer with weight decay, initial learning rate, batch size, and epoch as 0.3, 0.003, 64, and 350 respectively (where the learning rate is reduced by every 10 epochs after 110 epochs). In our experiments, we compare with 6 types of state-of-the-art methods, including DCRNN~\cite{li2018diffusion}, STGCN~\cite{yu2018spatio}, GraphWaveNet~\cite{wu2019graph}, AGCRN~\cite{bai2020adaptive}, Z-GCNETs~\cite{chen2021z}, and StemGNN~\cite{cao2020spectral}. We search the hidden feature dimension of the CNN-based model for TMP representation learning among $\{16, 32, 64, 128\}$, and the embedding dimension among values $\{1,2,3,5,10\}$. The resolution of TMP is 50 for all three datasets (i.e., the shape of input TMP is $50 \times 50$). The tuning of our proposed TMP-Nets on each dataset is done via grid search over a fixed set of choices and the same cross-validation setup is used to tune the above baselines. For both PeMSD4 and PeMSD8, specifically, we consider the first 1,000 and 2,000 timestamps for both of them. This allows us to further explore the learning capabilities of our Z-GCNETs as a function of sample size and also most importantly to assess the performance of Z-GCNETs and its competitors under a more challenging and much more realistic scenario of limited temporal records. For all methods (including our TMP-Nets and baselines), we run 5 times in the same partition and report the average accuracy along with the standard deviation. 


%

\subsubsection{Filtering Functions and Thresholds.} We select five filtering functions capturing different graph properties: 3 node sublevel filtrations (degree, closeness, and betweenness) and 2 power filtrations on edges (transaction and volume). The thresholds are chosen from equally spaced quantiles of either, function values (node sublevel filtration), or geodesic distances (power filtration). As a result, the number of thresholds depends upon the desired resolution for the MP grid. A too-low resolution does not provide sufficient topological information for graph learning tasks, whilst a too-high resolution unnecessarily increases the computational complexity. Based on our cross-validation experiments, we found 50 thresholds is a reasonable rule of thumb, working well in most studies.

\subsubsection{Prediction Horizon.} For Ethereum blockchain token networks (i.e., Bytom, Decentraland, and Golem), according to~\cite{chen2021z}, we set the forecasting step as 7 and the sliding window size as 7. For traffic datasets (PeMSD4 and PeMD8), according to~\cite{guo2019attention}, we set the forecasting step as 5 and the sliding window size as 12.



\section{More on Persistent Homology}
\label{Sec:MorePH}

\subsection{Zigzag Persistent Homology} \label{sec:zigzag}


While the notion of zigzag persistence is general, to keep the exposition simpler, we restrict ourselves to dynamic networks. For a given dynamic network $\wt{\G}=\{\mathcal{G}_t\}_{1}^T$, zigzag persistence detects pairwise compatible topological features in this time-ordered sequence of networks. While in single PH inclusions are always in the same direction (forward or backward), 
zigzag persistence and, more generally, the Mayer–Vietoris Diamond Principle allows us to consider
evolution of topological features
simultaneously in multiple directions~\cite{carlsson2009zigzag}.
In particular, define
a set of network inclusions over time


$$
\begin{matrix} \label{net_incl}
\mathcal{G}_1  & & & & \mathcal{G}_{2}& & & & \mathcal{G}_{3} & \ldots, \\
  & \sehookarrow & & \swhookarrow &  & \sehookarrow & & \swhookarrow & &  \\
  & & \mathcal{G}_1 \cup \mathcal{G}_{2} & &  & & \mathcal{G}_2 \cup \mathcal{G}_{3} & & &
\end{matrix}
$$
where $\mathcal{G}_k \cup \mathcal{G}_{k+1}$ is defined as a graph with a node set $V_k \cup V_{k+1}$ and an edge set $E_k \cup E_{k+1}$. 

Then, as before, by going to clique complexes of $\G_i$ and $\G_i\cup\G_{i+1}$, we obtain an "extended" zigzag filtration induced by the dynamic network $\{\mathcal{G}_t\}_{1}^T$, which allows us to detect the topological features which persist over time. 
That is, we record time points where we first and last observe a topological feature $\sigma$ over 
the considered time period,
i.e.,
birth and death times of $\sigma$, respectively, and $1\leq d_\sigma< b_\sigma \leq T$.
Notice that in contrast to ordinary persistence, both birth and death times ($b_\sigma$ or $d_\sigma$) can be fractional $i+\frac{1}{2}$ (corresponding to $\G_i\cup\G_{i+1}$) for $1\leq i<T$. 
We then obtain \textit{$k^{th}$ Zigzag Persistence Diagram}
${\rm{ZPD}_k}(\wt{\G})=\{(b_\sigma, d_\sigma) \mid \sigma\in H_k(\wh{\mathcal{G}}_i) \mbox{ for } b_\sigma\leq i<d_\sigma\}$.

\subsection{Stability for Single Persistence Vectorizations} \label{sec:stability2}
For a given PD vectorization, stability is one of the most important properties for statistical purposes. Intuitively, the stability question is whether a small perturbation in PD causes a big change in the vectorization or not. To make this question meaningful, one needs to formalize what "small" and ``big" means in this context. That is, we need to define a notion of distance, i.e., a metric in the space of PDs. The most common such metric is called
\textit{Wasserstein distance} (or matching distance) which is defined as follows. Let $PD(\X^+)$ and $PD(\X^-)$ be persistence diagrams two datasets $\X^+$ and $\X^-$. (We omit the dimensions in PDs). 
Let $PD(\X^+)=\{q_j^+\}\cup \Delta^+$ and  $PD(\X^-)=\{q_l^-\}\cup \Delta^-$ where 
$\Delta^\pm$ represents the diagonal (representing trivial cycles) with infinite multiplicity. Here, $q_j^+=(b^+_j,d_j^+)\in PD(\X^+)$ represents the birth and death times of a $k$-hole $\sigma_j$. 
Let $\phi:PD_k(\X^+)\to PD_k(\X^-)$ represent a bijection (matching). With the existence of the diagonal $\Delta^\pm$ on both sides, we make sure of the existence of these bijections even if the cardinalities $|\{q_j^+\}|$ and $|\{q_l^-\}|$ are different. Then, $p^{th}$ Wasserstein distance $\W_p$ defined as $$\W_p(PD(\X^+),PD(\X^-))= \min_{\phi}(\sum_j\|q_j^+-\phi(q_j^+)\|_\infty^p)^\frac{1}{p},$$
where $p\in \mathbb{Z}^+$. Here, the bottleneck distance is $\W_\infty(PD(\X^+),PD(\X^-))=\max_j \|q_j^+-\phi(q_j^+)\|_\infty$.

Then, function $\varphi$ is called \textit{stable} if $\mathrm{d}(\varphi^+,\varphi^-)\leq C\cdot \W_p(PD(\X^+),PD(\X^-))$, where $\varphi^\pm$ is a vectorization of $PD(\X^\pm)$ and $\mathrm{d}(.,.)$ is a suitable metric in the space of vectorizations. Here, the constant $C>0$ is independent of $\X^\pm$. This stability inequality interprets that as the changes in the vectorizations are bounded by the changes in PDs. 
If a given vectorization $\varphi$ holds such a stability inequality for some $\mathrm{d}$ and $\W_p$, we call $\varphi$ a \textit{stable vectorization}~\cite{atienza2020stability}. Persistence Landscapes~\cite{Bubenik:2015}, Persistence Images~\cite{adams2017persistence}, Stabilized Betti Curves~\cite{johnson2021instability} and several Persistence curves~\cite{chung2019persistence} are among well-known examples of stable vectorizations. 

\section{More on TMP Vectorizations} \label{sec:moreTMP}

\subsection{Further Examples of TMP Vectorizations} \label{sec:MPVexamples}

\subsubsection{TMP Silhouettes.} Silhouettes are another very popular SP vectorization method in machine learning applications~\cite{chazal2014stochastic}. The idea is similar to persistence landscapes, but this vectorization uses the life span of the topological features more effectively. For $PD(\G)=\{(b_i,d_i)\}_{i=1}^N$, let $\Lambda_i$ be the generating function for $(b_i,d_i)$ as defined in Landscapes (\cref{sec:TMP}). Then, \textit{Silhouette} function $\psi$ is defined as $\psi(\G)=\dfrac{\sum_{i=1}^N w_i\Lambda_i(t)}{\sum_{i=1}^m w_i}, \ t\in[\e_1,\e_q]$, where the weight $w_i$ is mostly chosen as the lifespan $d_i-b_i$, and $\{\e_k\}_{k=1}^q$ represents the thresholds for the filtration used. Again such a Silhouette function $\psi(\G)$ produces a $1D$-vector $\vec{\psi}(\G)$ of size $1\times (2q-1)$ as in persistence landscapes case.



As the structures of Silhouettes and Persistence Landscapes are very similar, so are their TMP Vectorizations.  For a given time-dependent data $\wt{\G}=\{\G_t\}_{t=1}^T$, similar to persistence landscapes, we use time direction and filtering function direction for our TMP Silhouettes. For a filtering function $f:\V_t\to \R$ with threshold set $\I=\{\alpha_j\}_1^m$, we obtain \textit{TMP Silhouette} as $\M_\psi^j(\wt{\G})=\vec{\psi}(\wt{\G}^j)$, where $\M_\psi^j$ represents $j^{th}$-row of the $2D$-vector $\M_\psi$ and $\vec{\psi}(\wt{\G}^j)$ is the Silhouette vector induced by the zigzag persistence diagram for the time sequence $\wt{\G}^j$. Again similar to the landscapes, by taking $q=2T-1$, $\M_\psi(\wt{\G})$, we obtain a $2D$-vector of size $m\times (4T-3)$, where $T$ is the number of time steps in the data $\wt{\G}$.

\subsubsection{TMP Betti \& TMP  Persistence Summaries.}
Next, we discuss an important family of SP vectorizations, Persistence Curves~\cite{chung2019persistence}. This is an umbrella term for several different SP vectorizations, i.e. Betti Curves, Life Entropy, Landscapes, et al. Our TMP framework naturally adapts to all Persistence Curves to produce multidimensional vectorizations. As Persistence Curves produce a single variable function in general, they all can be represented as 1D-vectors by choosing a suitable mesh size depending on the number of thresholds used. Here, we describe one of the most common Persistence Curves in detail, i.e. Betti Curves. It is straightforward to generalize the construction to other Persistence Curves.

\begin{figure*}[t]
    \centering
    \includegraphics[width=0.9\textwidth]{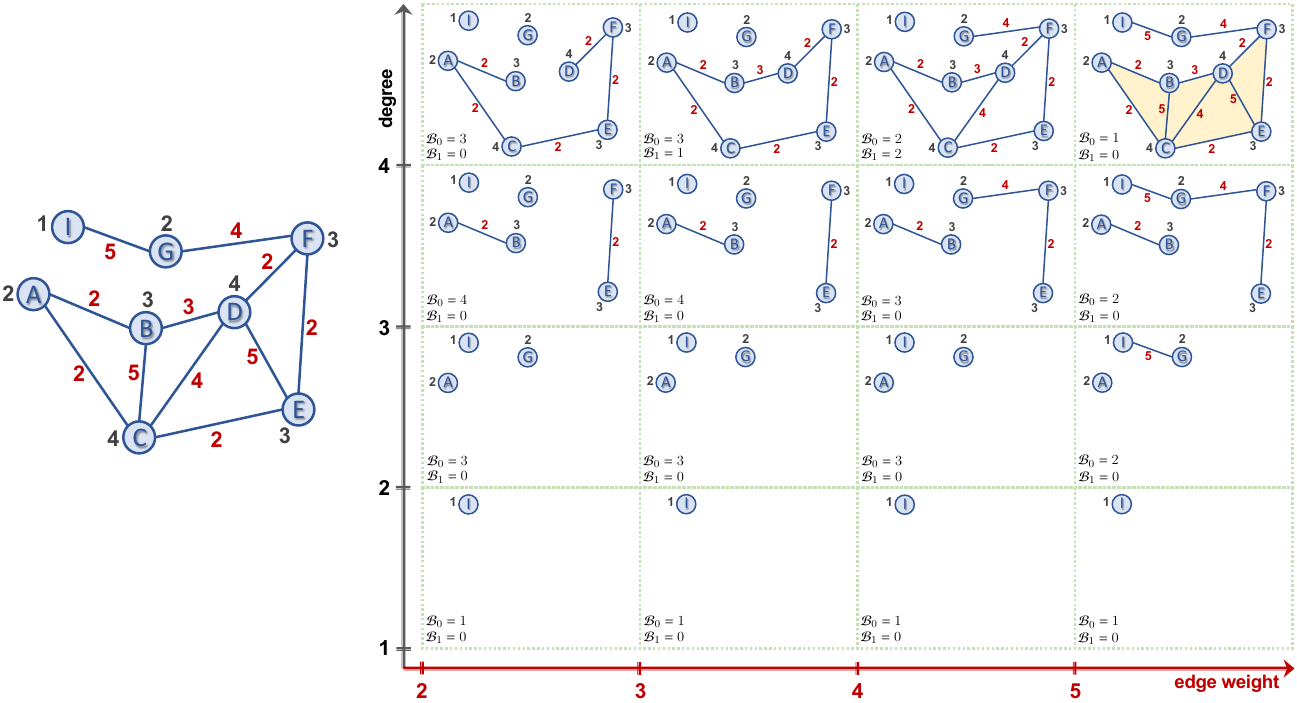}
    \caption{\scriptsize Multidimensional persistence on a graph network (original graph: left). Black numbers denote the degree values of each node whilst red numbers show the edge weights of the network. Hence, shape properties are computed on two filtering functions (i.e., degree and edge weight). While each row filters by degree, each column filters the corresponding subgraph using its edge weights. For each cell, lower left corners represent the corresponding threshold values. For each cell, $\mathcal{B}_{0}$ and $\mathcal{B}_{1}$ represent the corresponding Betti numbers.  \label{Fig:ToyMP}}  
\end{figure*}

Betti curves are one of the simplest SP vectorizations as it gives the count of the topological feature at a given threshold interval. In particular, $\beta_k(\Delta)$ is the total count of $k$-dimensional topological feature in the simplicial complex $\Delta$, i.e. $\beta_k(\Delta)=rank(H_k(\Delta))$ (See \cref{Fig:ToyMP}). For a given time-dependent data $\wt{\G}=\{\G_t\}_{t=1}^T$, we use zigzag persistence in time direction, which yields $2T-1$ thresholds steps. Then, the \textit{Betti Curve} $\beta(\wt{\G})$ for the zigzag persistence diagram $ZPD(\wt{\G})$ is a step function with $2T-1$-intervals (we add another threshold after $2T$ to interpret the last interval). As $\beta(\wt{\G})$ is a step function, it can be described as a vector of size $1\times (2T-1)$, i.e.  $\vec{\beta}(\wt{\G})=[ \beta(1)\ \beta(1.5)\ \beta(2)\ \beta(2.5)\ \beta(3)\ \dots \ \beta(T)]$, where $\beta(t)$ is the total count of topological features in $\wh{\G}_t$. 
Here we omit the homological dimensions (i.e., subscript $k$) to keep the exposition simpler. 

Then, by using a filtering function $f:\V_t\to \R$ with threshold set $\I=\{\alpha_j\}_1^m$ for other direction, we define a \textit{TMP Betti curve} as $\M_\beta^j=\vec{\beta}(\wt{\G}^j)$, where $\M_\beta^j$ is the $j^{th}$-row of the $2D$-vector $\M_\beta$. Here, $\wt{\G}^j=\{\G_t^j\}_{t=1}^T$ is induced by the sublevel filtration for $f:\V_t\to\R$, i.e. $\V^j_t=\{v_r\in \V_t\mid f(v_r)\leq \alpha_j\}$.  Then, $\M_\beta$ is s $2D$-vector of size $m\times (2T-1)$.

An alternate (and computationally friendly) route for TMP Betti Summaries is to bypass zigzag persistent homology and $\wh{\G}_{t+\frac{1}{2}}$ cliques and use directly clique complexes $\{\wh{\G}_t\}_{t=1}^T$. This is because Betti curves do not need PDs, and they can be directly computed from the simplicial complexes $\{\wh{\G}_t\}_{t=1}^T$. This way, we obtain a vector of size $1\times T$ as $\vec{\beta}(\wt{\G})=[ \beta(1)\ \beta(2)\ \beta(3)\ \dots \ \beta(T)]$. Then, this version of induced TMP Betti curve $\M_\beta(\wt{\G})$ yields a $2D$-vector of size $m\times T$. It might have less information than the original zigzag version, but this is computationally much faster~\cite{lesnick2022computing} as one skips computation of PDs. 
Note that skipping zigzag persistence in time direction is only possible for Betti curves, as other vectorizations come from PDs, that is, lifespans, birth and death times are needed.

\subsection{TMP Vectorizations and Multipersistence} \label{sec:MP_theory}

Multipersistence theory is under intense research because of its promise to significantly improve the performance and robustness properties of single persistence theory. While single persistence theory obtains the topological fingerprint of single filtration, a multidimensional filtration with more than one parameter should deliver a much finer summary of the data to be used with ML models. However, multipersistence virtually has not reached any applications yet and remains largely unexplored by the ML community because of technical problems. Here, we provide a short summary of these issues. For further details, \cite{botnan2022introduction}~gives a nice outline of the current state of the theory and major obstacles.

In single persistence, the threshold space $\{\alpha_i\}$ being a subset of $\mathbb{R}$, is totally ordered, i.e., birth time~$<$~death time for any topological feature appearing in the filtration sequence $\{\Delta_i\}$. By using this property, it was shown that “barcode decomposition” is well-defined in single persistence theory in the 1950s [Krull-Schmidt-Azumaya Theorem~\cite{botnan2022introduction}--Theorem 4.2]. This decomposition makes the persistence module $M=\{H_k(\Delta_i)\}_{i=1}^N$ uniquely decomposable into barcodes. This barcode decomposition is exactly what we call a PD.

However, when one goes to higher dimensions, i.e. $d=2$, then the threshold set $\{(\alpha_i,\beta_j)\}$ is no longer totally ordered, but becomes partially ordered (Poset). In other words, some indices have ordering relation $(1,2)< (4,7)$, while some do not, e.g., (2,3) vs. (1,5). Hence, if one has a multipersistence grid $\{\Delta_{ij}\}$, 
we no longer can talk about birth time or death time as there is no ordering any more. Furthermore, Krull-Schmidt-Azumaya Theorem is no longer true for upper dimensions~\cite{botnan2022introduction}--Section 4.2. Hence, for general multipersistence modules barcode decomposition is not possible, and the direct generalization of single persistence to multipersistence fails. On the other hand, even if the multipersistence module has a good barcode decomposition, because of partial ordering, representing these barcodes faithfully is another major problem. Multipersistence modules are an important subject in commutative algebra, where one can find the details of the topic in~\cite{eisenbud2013commutative}. 

While complete generalization is out of reach for now, several attempts have been tried to utilize the MP idea by using one-dimensional slices in the MP grid in recent~\cite{carriere2020multiparameter,vipond2020multiparameter}.  Slicing techniques use the persistence diagrams of predetermined one-dimensional slices in the multipersistence grid and then combine (compress) them as one-dimensional output~\cite{botnan2022introduction}. In that respect, one major issue is that the topological summary highly depends on the predetermined slicing directions in this approach. The other problem is the loss of information when compressing the information in various persistence diagrams.



As explained above, the MP approach does not have theoretical foundations yet, and there are several attempts to utilize this idea. In this paper, we do not claim to solve theoretical problems of multipersistence homology but offer a novel, highly practical multidimensional topological summary by advancing the existing methods in spatio-temporal settings. We use the time direction in the multipersistence grid as a natural slicing direction and overcome the predetermined slices problem. Furthermore, for each filtering step of the spatial direction, unlike other MP vectorizations, we do not compress the induced PDs, but we combine them as multidimensional vectors (matrices or arrays). As a result, these multidimensional topological fingerprints are capable of capturing very fine topological information hidden in the spatio-temporal data. In the spatial direction, we filter the data with one (or more) domain function and obtain induced substructures, while in the time direction, we capture the evolving topological patterns of these substructures induced by the filtering function via zigzag persistence. Our fingerprinting process is highly flexible, one can easily choose the right single persistence vectorization to emphasize either density of short barcodes or give importance to long barcodes appearing in these PDs. We obtain multidimensional vectors (matrices and arrays) as output which are highly practical to be used with various ML models.

\subsection{TMP Framework for General Types of Data} \label{sec:othertypedata}


So far, to keep the exposition simpler, we described our construction for dynamic networks. However, our framework is suitable for various types of time-dependent data. Let $\wt{\X}=\{\X_t\}_{t=1}^T$ be a time sequence of images or point clouds. Let $f:\wt{\X}\to \R$ be a filtering function that can be applied to all $\X_t$. Ideally, $f$ is a function that does not depend on $t$, e.g if $\{\X_t\}$ represent a sequence of images for time $1\leq t \leq T$, $f$ can be taken as a grayscale function. If $\{\X_t\}$ is a sequence of point clouds at different times, then $f$ can be defined as a density function.

Then, the construction is the same as before. Let $f:\wt{\X}\to \R$ be the filtering function with threshold set $\I=\{\alpha_j\}_1^m$. Let $\X_t^j=f^{-1}((-\infty,\alpha_j])$. Then, for each $1\leq j_0\leq m$, we have a time sequence $\wt{\X}^{j_0}=\{\X_t^{j_0}\}_{t=1}^T$. Let  $\{\wh{\X}_t^{j_0}\}_{t=1}^T$ be the induced simplicial complexes to be used for filtration. Then, by taking $\wh{\X}^{j_0}_{k.5}=\wh{\X}^{j_0}_{k}\cup\wh{\X}^{j_0}_{k+1}$, we apply Zigzag PH to this sequence as before.
$$\wh{\X}_1^{j_0}\hookrightarrow\wh{\X}^{j_0}_{1.5}\hookleftarrow\wh{\X}^{j_0}_2\hookrightarrow\wh{\X}^{j_0}_{2.5}\hookleftarrow\wh{\X}^{j_0}_3\hookrightarrow\dots\hookleftarrow\wh{\X}^{j_0}_T$$

Then, we obtain the zigzag persistence diagram $ZPD(\wt{\X}^{j_0})$ for the filtration $\{\wh{\X}_t^{j_0}\}_{t=1}^T$. Hence, we obtain $m$ zigzag PDs, $ZPD(\wt{\X}^j)$, one for each $1\leq j\leq m$. Then, again by applying a preferred SP vectorization $\varphi$ to the persistence diagram $ZPD(\wt{\X}^j)$, we have corresponding vector $\vec{\varphi}(\wt{\X}^j)$ (say size $1\times k$). Then, TMP vectorization $\M_\varphi$ can be defined as $\M_\varphi^j(\wt{\X})=\vec{\varphi}(\wt{\X}^j)$ where $\M_\varphi^j$ represents $j^{th}$-row of the $2D$-vector $\M_\varphi$. Hence, TMP vectorization of $\wt{\X}$, $\M_\varphi(\wt{\X})$, becomes a $2D$-vector of size $m\times k$.

\section{Stability of TMP Vectorizations}  \label{sec:stability3}

In this part, we prove the stability theorem (Theorem~\ref{thm:stability}) for TMP vectorizations. In particular, we prove that if the original SP vectorization $\varphi$ is stable, then so is its TMP vectorization $\M_\varphi$. Let $\wt{\G}=\{\G_t\}_{t=1}^T$ and $\wt{\h}=\{\h_t\}_{t=1}^T$ be two time sequences of networks. Let $\varphi$ be a stable SP vectorization with the stability equation  
\begin{equation}\label{eqn1}
\mathrm{d}(\varphi(\wt{\G}),\varphi(\wt{\h}))\leq C_\varphi\cdot \W_{p_\varphi}(PD(\wt{\G}),PD(\wt{\h}))
\end{equation}
for some $1\leq p_\varphi\leq \infty$. Here, $\W_p$ represents Wasserstein-$p$ distance as defined before. 

Now, by taking $d=2$ for TMP construction, we obtain bifiltrations $\{\wh{\G}_t^{ij}\}$ and $\{\wh{\h}_t^{ij}\}$ for each $1\leq t\leq T$. We define the induced matching distance between the multiple PDs as 
\begin{equation}\label{eqn2}
 \resizebox{\columnwidth}{!}{$\displaystyle \mathbf{D}(\{ZPD(\wt{\G})\},\{ZPD(\wt{\h})\})=\max_{i,j}\W_{p_\varphi}(ZPD(\wt{\G}^{ij}), ZPD(\wt{\h}^{ij}))$}
\end{equation}

Now, we define the distance between induced TMP Vectorizations as 
\begin{equation}\label{eqn3}
\mathbf{D}(\M_\varphi(\wt{\G}),\M_\varphi(\wt{\h}))=\max_{i,j} \mathrm{d}(\varphi(\wt{\G}^{ij}),\varphi(\wt{\h}^{ij})).
\end{equation}



\noindent {\bf Theorem~\ref{thm:stability}:} {\em Let $\varphi$ be a stable vectorization for single parameter PDs. Then, the induced TMP Vectorization $\M_\varphi$ is also stable, i.e. 
there exists $\wh{C}_\varphi>0$ such that for any pair of time-aware network sequences $\wt{\G}$ and $\wt{\h}$, the following inequality holds}
$$\mathbf{D}(\M_\varphi(\wt{\G}),\M_\varphi(\wt{H}))\leq \wh{C}_\varphi\cdot \mathbf{D}(\{ZPD(\wt{\G})\},\{ZPD(\wt{\h})\}).$$

\begin{proof} WLOG, we assume SP vectorization $\varphi$ produces a $1D$-vector. For $2D$ or higher dimensional vectors, the proof will be similar. For any $i_0,j_0$ with $1\leq i_0\leq m$ and $1\leq j_0\leq n$, we will have filtration sequences $\{\wh{\G}^{i_0j_0}_t\}_{t=1}^T$ and $\{\wh{\h}^{i_0j_0}_t\}_{t=1}^T$. This produces a zigzag persistence diagrams $ZPD(\wt{\G}^{i_0 j_0})$ and $ZPD(\wt{\h}^{i_0 j_0})$. Therefore, we have $m.n$ pairs of zigzag persistence diagrams. 

Consider the distance definition for TMP vectorizations (\cref{eqn3}). Let $i_1, j_1$ be the indices realizing the maximum in the right side of the equation, i.e. 
\begin{equation}\label{eqn4}
\mathrm{d}(\varphi(\wt{\G}^{i_1j_1}),\varphi(\wt{\h}^{i_1j_1}))=\max_{i,j} \mathrm{d}(\varphi(\wt{\G}^{ij}),\varphi(\wt{\h}^{ij})).
\end{equation}
Then, by stability of $\varphi$ (i.e., inequality (\ref{eqn1})), we have 
\begin{equation}
\label{eqn5} \resizebox{\columnwidth}{!}{
$\mathrm{d}(\varphi(\wt{G}^{i_1j_1}),\varphi(\wt{\h}^{i_1j_1})) \leq C_\varphi\cdot \W_{p_\varphi}(ZPD(\wt{G}^{i_1j_1}),ZPD(\wt{\h}^{i_1j_1}))$.}
\end{equation}

Now, as 
\begin{equation} \resizebox{\columnwidth}{!}{
$\W_{p_\varphi}(ZPD(\wt{G}^{i_1j_1}),ZPD(\wt{\h}^{i_1j_1})) \\ \leq \max_{i,j}\W_{p_\varphi}(ZPD(\wt{\G}^{ij}), ZPD(\wt{\h}^{ij}))$,}
\end{equation} 
by the definition of distance between TMP vectorizations (\cref{eqn2}), we find that 
\begin{equation} \resizebox{\columnwidth}{!}{ 
$\W_{p_\varphi}(ZPD(\wt{G}^{i_1j_1}),ZPD(\wt{\h}^{i_1j_1})) \\ \leq \mathbf{D}(\{ZPD(\wt{\G})\},\{ZPD(\wt{\h})\})$.}
\end{equation}
Finally,   
\begin{eqnarray*}
\mathbf{D}(\M_\varphi(\wt{\G}),\M_\varphi(\wt{\h}))&=&\mathrm{d}(\varphi(\wt{\G}^{i_1j_1}),\varphi(\wt{\h}^{i_1j_1}))\\ &\leq &  C_\varphi\cdot \W_{p_\varphi}\\ &\leq & \wh{C}_\varphi\cdot \mathbf{D}(\{ZPD(\wt{\G})\},\{ZPD(\wt{\h})\}). \nonumber
\end{eqnarray*}
Here, the leftmost equation follows from~\cref{eqn3}. The first inequality follows from~\cref{eqn5}. The final inequality follows from~\cref{eqn2}. This concludes the proof of the theorem.
\end{proof}

\begin{remark} [Stability with respect to Matching Distance] \label{rmk:matching_distance} While we define our own distance $\mathbf(.,.)$ in the space of MP modules for suitability to our setup, if you take $\textit{matching\ distance}$ in the space of MP modules, our result still implies the stability for TMP vectorizations $\mathbf{M}_{\varphi}$ induced from stable SP vectorization $\varphi$ with $p_\varphi=\infty$. In particular, our distance definition $\mathbf{D}(.,.)$ with specified slices is just a version of matching distance $d_M(.,.)$  restricted only to horizontal slices. The matching distance between two multipersistence modules $E_1, E_2$ is defined as the supremum of the bottleneck ($W_\infty(.,.)$) distances between single persistence diagrams induced from all one-dimensional fibers (slices) of the MP module, i.e. $d_M(E_1,E_2)=\sup_L W_\infty(PD(E_1(L), PD(E_2(L)))$ where $E_i(L)$ represents the slice restricted to line $L$ in the MP grid $E_i$ \cite[Section 12.3]{dey2022computational}. 

In this sense, with this notation, our distance would be a restricted version of matching distance $d_M(.,.)$ as follows: $\mathbf{D}(M_1,M_2)=\max W_\infty,$ where $L_0$ represent horizontal slices, then $\mathbf{D}(E_1,E_2)\leq  d_M(E_1,E_2)$. Then, by combining with our stability theorem, we obtain $d(\mathbf{M}_\varphi(E_1),\mathbf{M}_\varphi(E_2))\leq C_\varphi d_M(E_1,E_2)$. Hence, if two MP modules are close to each other in the matching distance, then their corresponding TMP vectorizations $\mathbf{M}_\varphi$ are close to each other, too. 

To sum up, for TMP vectorizations $\mathbf{M}_{\varphi}$ induced from stable SP vectorization $\varphi$ with $p_\varphi=\infty$, our result naturally implies stability with respect to matching distance on multipersistence modules. The condition $p_\varphi=\infty$ comes from the bottleneck distance ($W_\infty$) used in the definition of $d_M$. If one defines a generalization of matching distance for other Wasserstein distances $W_p$ for $p\in [1,\infty)$, then a similar result can hold for other stable TMP vectorizations. 
\end{remark}

\begin{table}[t!]
\centering
\caption{Notation and main symbols.\label{notations}}
\begin{tabular}{ll}
\toprule
\textbf{Notation} &  \textbf{Definition} \\
\hline
$\mathcal{G}_t$ & the spatial network at timestamp $t$\\
$\mathcal{V}_t$ & the node set at timestamp $t$\\
$\mathcal{E}_t$ & the edge set at timestamp $t$\\
$W_t$ & the edge weights at timestamp $t$\\
$N_t$ & the number of nodes at timestamp $t$\\
$\wt{\G}$ & a time series of graphs\\
$\wh{\mathcal{G}}^{i}$ & an abstract simplicial complex\\
$D$ & the highest dimension in a simplicial complex\\
$\sigma$ & a $k$-dimensional topological feature\\
$d_\sigma-b_\sigma$ & the life span of $\sigma$\\
${\rm{PD}_k}(\G)$ & $k-$dimensional persistent diagram\\
$H_k(\cdot)$ & $k^{th}$ homology group\\
$f$ and $g$ & two filtering functions for sublevel filtration\\
$F(\cdot,\cdot)$ & multivariate filtering function\\
$m\times n$ & rectangular grid for bifiltrations\\
$ZPD_k(\cdot)$ & the zigzag persistence diagram\\
$\varphi$ & a single persistence vectorization\\
$\vec{\varphi}(\cdot)$ & the vector from $\varphi$\\
$\M_\varphi$ & TMP Vectorization of $\varphi$\\
$\lambda$ & persistence landscape\\
$\M_\lambda$ & TMP Landscape\\
$\wt{\mu}$ & persistence surface\\
$\M^j_\mu$ & TMP Persistence Image\\
$\mathbf{D}(\cdot, \cdot)$ & distance between persistence diagrams\\
$\mathrm{D}(\cdot, \cdot)$ & distance between TMP Vectorizations\\
\midrule
$X$ & node feature matrix\\
$Z^{(\ell)}_{t, \text{Spatial}}$ & graph convolution on adptive adjacency matrix\\
$Z_{t, \text{TMP}}$ & image-level local topological feature\\
$\W_p(\cdot, \cdot)$ & Wasserstein distance\\
$\W_\infty(\cdot, \cdot)$ & Bottleneck distance\\  
\bottomrule
\end{tabular}
\end{table}

\end{document}